\documentclass[lettersize,journal]{IEEEtran}
\usepackage{amsmath,amsfonts}
\usepackage{algorithmic}
\usepackage{array}
\usepackage[caption=false,font=normalsize,labelfont=sf,textfont=sf]{subfig}
\usepackage{textcomp}
\usepackage{stfloats}
\usepackage{url}
\usepackage{verbatim}
\usepackage{graphicx}
\usepackage{arydshln} 
\def\BibTeX{{\rm B\kern-.05em{\sc i\kern-.025em b}\kern-.08em
    T\kern-.1667em\lower.7ex\hbox{E}\kern-.125emX}}
\usepackage{balance}
\begin{document}
\title{3D-SeqMOS: A Novel Sequential 3D Moving Object Segmentation in Autonomous Driving}
\author{Qipeng Li, Yuan Zhuang,~\IEEEmembership{Member,~IEEE,} Yiwen Chen, Jianzhu Huai,~\IEEEmembership{Member,~IEEE,} \\
 Miao Li, Tianbing Ma, Yufei Tang and Xinlian Liang,~\IEEEmembership{Senior Member,~IEEE,}
\thanks{This work was supported by Excellent Youth Foundation of Hubei Scientific Committee 2021CFA040, GuangDong Basic Applied and Basic Research Foundation 2021A1515110343, Special Fund of Anhui University of Science and Technology SKLMRDPC21KF20. \emph{(Corresponding author: Yuan Zhuang)}

Qipeng Li, Yuan Zhuang, Yiwen Chen, Jianzhu Huai, Xinlian Liang are with the State Key Laboratory of Information Engineering in Surveying, Mapping and Remote Sensing, Wuhan University, Wuhan 430206, China (e-mail: qipeng\_li@163.com, yuan.zhuang@whu.edu.cn, cyw9620@gmail.com, huai.3@osu.edu, xinlian\_liang@hotmail.com). 

Miao Li is with the Institute of Technological Sciences, Wuhan University, Wuhan 430206, China (e-mail: miao.li@whu.edu.cn).

Tianbing Ma is with the Anhui University of Science and Technology, Huainan 232001, Anhui, China (e-mail: dfmtb@163.com).


Yufei Tang is with the Department of Computer Electrical Engineering and Computer Science, Florida Atlantic University, Boca Raton, FL 33431 USA (e-mail: tangy@fau.edu).

}}

\markboth{ }
{How to Use the IEEEtran \LaTeX \ Templates}

\maketitle

\begin{abstract}
For the SLAM system in robotics and autonomous driving, the accuracy of front-end odometry and back-end loop-closure detection determine the whole intelligent system performance. But the LiDAR-SLAM could be disturbed by current scene moving objects, resulting in drift errors and even loop-closure failure. Thus, the ability to detect and segment moving objects is essential for high-precision positioning and building a consistent map. In this paper, we address the problem of moving object segmentation from 3D LiDAR scans to improve the odometry and loop-closure accuracy of SLAM. We propose a novel 3D Sequential Moving-Object-Segmentation (3D-SeqMOS) method that can accurately segment the scene into moving and static objects, such as moving and static cars. Different from the existing projected-image method, we process the raw 3D point cloud and build a 3D convolution neural network for MOS task. In addition, to make full use of the spatio-temporal information of point cloud, we propose a point cloud residual mechanism using the spatial features of current scan and the temporal features of previous residual scans. Besides, we build a complete SLAM framework to verify the effectiveness and accuracy of 3D-SeqMOS. Experiments on SemanticKITTI dataset show that our proposed 3D-SeqMOS method can effectively detect moving objects and improve the accuracy of LiDAR odometry and loop-closure detection. The test results show our 3D-SeqMOS outperforms the state-of-the-art method by 12.4\emph{\%}. We extend the proposed method to the \emph{SemanticKITTI: Moving Object Segmentation} competition and achieve the \emph{2nd} in the leaderboard, showing its effectiveness.

\end{abstract}

\section{Introduction}
\IEEEPARstart{S}{imultaneous} Localization and Mapping (SLAM) is one of the basic functions of robotics or autonomous driv-\\ing. For a stable SLAM system, globally consistent mapping is the basis for accurate positioning in the environment. 3D LiDAR sensors can capture more precise and further-away distance measurements of the surrounding environments than conventional visual cameras \cite{li2020deep}. In the complex scene, such as dark light or lack of texture, although the LiDAR-SLAM has better robustness and stability than Visual-SLAM, it would be disturbed by the moving objects, resulting in the failure of odometry estimation or loop-closure detection, which are essential for mapping and positioning.

\begin{figure}[t]
\centering
\subfloat[]{\includegraphics[scale=0.13]{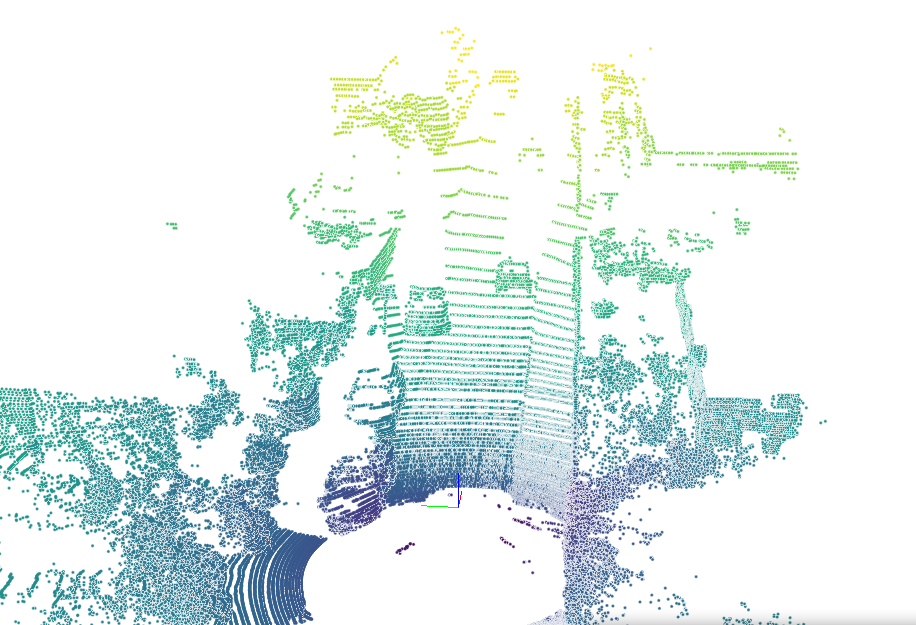}%
\label{fig_first_case}}
\hfil
\subfloat[]{\includegraphics[scale=0.13]{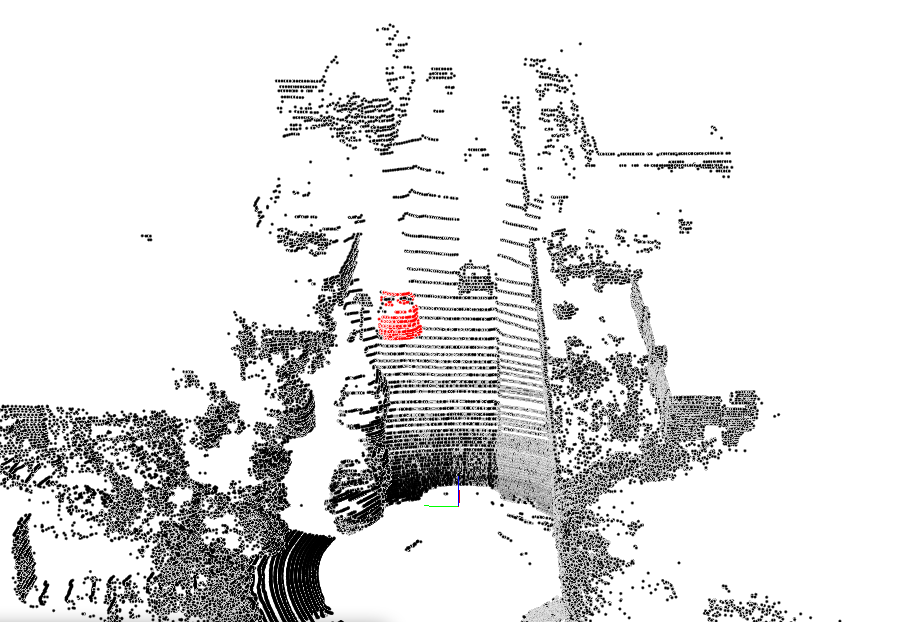}%
\label{fig_second_case}}
\hfil
\subfloat[]{\includegraphics[scale=0.25]{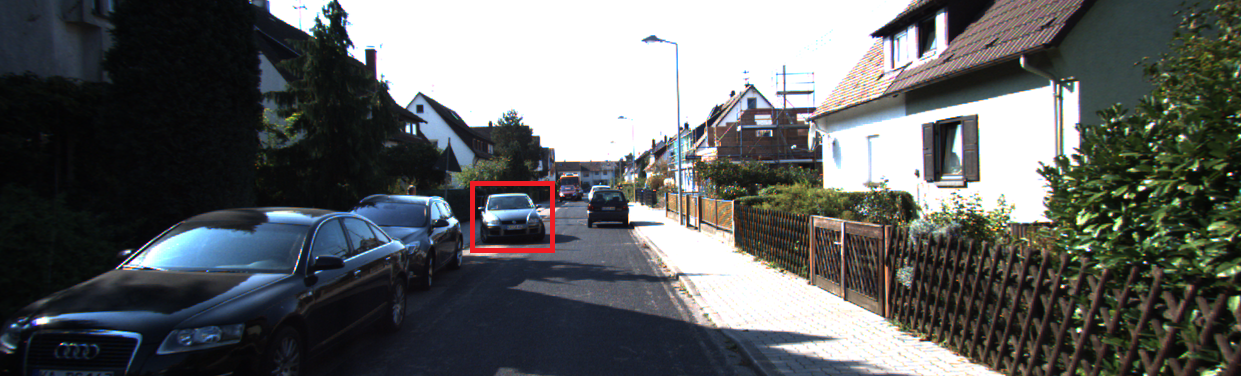}%
\label{fig_second_case}}
\caption{The moving object segmentation with our 3D-SeqMOS. Our proposed 3D-SeqMOS can segment moving objects using the sequential point clouds. (a) is the raw point cloud from SemanticKITTI dataset, (b) is the segmentation result of our method, and (c) is the color image of the current scene. In particular, we do not segment all movable objects, but only the current moving objects. As shown in (b), the red car is moving, while the car parked on both sides of the road is stationary.}
\label{fig_1}
\end{figure}

\begin{figure}[t]
\centering
\subfloat{\includegraphics[width=3.4in]{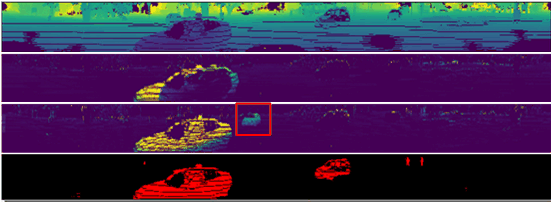}%
\label{image3}}
\caption{Using the spatial difference of adjacent frame point cloud to detect the result of moving target. Due to the large noise of point cloud scene, it cannot achieve reliable results. The four images from top to bottom are: 1) the original point cloud scene, 2) the spatial difference result between current frame and previous frame, 3) the spatial difference between current frame and previous second, i.e. the previous 10-th frame, and 4) the ground truth of the moving objects in current point cloud scene, marked in red.}
\label{fig_1.2}
\end{figure}

When there are a large number of moving objects, they bring disturbances to the robotics or autonomous driving SLAM system, especially to the front-end Lidar-SLAM odometry, resulting in large position drift error of the system. Therefore, identifying moving parts from static scene is crucial for safe and reliable autonomous positioning and navigation. In our work, we propose a moving object segmentation method which uses raw 3D point cloud to recognize moving and static objects, as shown in Fig. \ref{fig_1}. As a result, autonomous driving can effectively use the point cloud information in the scene to identify the moving object, predict its future state, avoid collision, and optimize the trajectory planning. In order to solve the above moving object segmentation (MOS) problem, many scholars have proposed corresponding solutions. Inspired by the traditional image processing methods, although the point cloud can be transformed into 2D image to calculate the spatial difference for the purpose of detecting moving objects, the complexity of the point cloud scene with large noise makes it difficult to segment the moving point cloud effectively, as shown in Fig. \ref{fig_1.2}. As shown in the second figure, it can only extract the boundary instead of the complete area. At the same time, the performance depends on the threshold between frames for different motion objects, such as cars or pedestrians, even recognize the same object as different targets, as shown in the red box of the third figure. When there is shelter or relative movement with the same speed, the depth value of the target changes slightly. Therefore, we adopt the learning-based approach for our research. With the rapid development of deep learning, the accuracy of image instance segmentation or detection based on semantic information has been high. However, in the research field of point cloud, due to the non-uniformity of its density distribution and disorder, moving object segmentation performance still needs to be improved.


For moving categories, the existing point cloud semantic segmentation methods are difficult to distinguish whether movable object is static or moving, such as a stopping car or a moving car. Besides, many methods convert 3D point clouds into 2D images in data preprocessing, and then process semantic information with the help of computer vision technology. LiDAR-MOS \cite{chen2021moving} is an advanced method at present, which has achieved excellent moving object segmentation performance. It converts 3D point clouds into 2D projected-images, and then use image-based semantic segmentation method to segment moving objects. However, this will lose the raw 3D structure and topology information of the point clouds, which makes it difficult for the image-based point cloud moving object segmentation method to achieve high accuracy. our work try to use the raw 3D point cloud to solve the problem of moving object segmentation without projected-images. In addition, many 3D point cloud neural network structures take single scan as the input \cite{zhu2021cylindrical}, and cannot deal with continuous multi-scans to use the temporal information. However, the temporal information will be more conducive to detect and segment moving objects. Our work also tries to develop a residual mechanism to extract the temporal features, and then combine the temporal-spatial information to train the network structure.

In this paper, we propose a novel 3D convolutional neural network for moving object segmentation of 3D point cloud. It should be noted that we process the raw 3D point cloud instead of converting the point cloud into an image. On the SemanticKITTI dataset, we verify the effectiveness and accuracy of our 3D-SeqMOS method, achieving the state-of-the-art (\emph{SOTA}) performance. In addition, to make full use of the temporal information provided by previous sequential scans, we propose a point cloud residual mechanism, and use the spatial information of the current scan and the temporal information of previous multi-scans to train our proposed network model. The original intention of designing 3D-SeqMOS is to improve the positioning and mapping accuracy of autonomous vehicles in complex scenes, so we proposed a complete SLAM framework, including front-end odometry and back-end loop-closure detection. Finally, a large number of experiments show that our 3D-SeqMOS effectively improves the positioning, mapping and loop-closure performance of SLAM.

The main contributions of our work are summarized:
\begin{itemize}
\item[1)]
In order to reduce the influence of moving objects on mapping and positioning, we propose a novel 3D point cloud moving object segmentation method (3D-SeqMOS), which achieves better segmentation accuracy than many existing state-of-the-art methods. Besides, we adopt a new loss function with voxel loss and point loss. We show that the direct process of the point cloud without projected-image is essential for moving object segmentation. 

\item[2)]
We propose a point cloud residual mechanism, which make full use of temporal information of previous multi-scans to represent the motion features. The spatial information of current scan and the temporal information of previous multi-scans are fused in our moving object segmentation network.

\item[3)]
In order to verify the effectiveness of our 3D-SeqMOS in autonomous driving, we propose a robust and accurate LiDAR-SLAM system, including front-end odometry and back-end loop-closure detection. With the proposed 3D-SeqMOS, the location accuracy of SLAM achieves the state-of-the-art performance. In addition, our 3D-SeqMOS can be perfectly combined with other existing SLAM methods to improve the accuracy and robustness, including odometry, mapping and loop-closure detection.

\end{itemize}

The remainder of this article is organized as follows. Section II focuses on related works. In Section III, the whole network architecture of proposed 3D-SeqMOS and the detail of each important module are provided. In Section IV, the experimental results are presented, which include moving object segmentation performance on challenging dataset and the improvement effect of our method on SLAM in robotics or autonomous driving. In Section V, this work is summarized, and the practicability and compatibility of our method are illustrated.

\section{Related Works}

\subsection{Moving Object Segmentation for Point Clouds}

Kim \emph{et al}. \cite{kim2020remove} proposed a method based on distance image, which used the consistency check between query scanning and pre-constructed map to remove dynamic points, and used multi-resolution error prediction restoration algorithm to refine the map. Although this map-based method can separate moving object from the background, they need pre-built and cleaned maps, so they usually cannot be operated online. At present, the semantic segmentation method based on LiDAR point cloud has achieved great success  \cite{zhu2021cylindrical,hu2020randla,cortinhal2020salsanext,li2021rethinking, du2022novel, corcoran2011background}. Semantic segmentation can be regarded as a related step of moving object segmentation. However, most existing semantic segmentation methods can only find movable object, such as vehicles and people, but cannot distinguish between actually moving object, such as moving cars or walking pedestrians, and non-moving/stationary object, such as parked cars or building structures. Some existing LiDAR moving object segmentation methods projected the 3D point cloud into the 2D space, and processed it through the 2D convolution network, such as  \cite{alonso20203d,milioto2019rangenet++}, based on the distance image or based on the aerial view image. However, these set of methods lose the raw 3D structure and topology information in the process of 3D to 2D projection. A natural alternative is to use 3D partitioning and 3D convolution networks to process point cloud and maintain their 3D geometric relationships. As for outdoor scene point cloud segmentation, most existing point cloud segmentation methods \cite{hu2020randla,cortinhal2020salsanext,milioto2019rangenet++,zhang2020deep} focused on converting 3D point cloud into 2D mesh, so that 2D convolutional neural network can be used. Squeezeseg \cite{wu2018squeezeseg}, Darknet  \cite{behley2019semantickitti}, Squeezesegv2 \cite{wu2019squeezesegv2} and Rangenet++ \cite{milioto2019rangenet++} used spherical projection mechanism to convert point cloud into front view image or distance image, and used 2D convolution network on pseudo image for point cloud segmentation. Yoon \emph{et al}. \cite{yoon2019mapless} used heuristics, such as residuals between LiDAR scanning, free space inspection and region growth, to detect moving objects. LiDAR-MOS \cite{chen2021moving} obtained the excellent moving object segmentation performance, which does not rely on pre-built maps, that is, only using the past LiDAR scanning, but operates by converting 3D point cloud to 2D image. We use the residual between the current frame and the previous frames as the additional input of the semantic segmentation network to realize category independent moving object segmentation. 

Our work is no longer to convert 3D point cloud into 2D image, but to process the raw point cloud directly. On the basis of cylinder partition, we improve the network structure of point cloud segmentation, so that we can directly segment the moving objects of point cloud in the current scene, and improve the performance compared with the based-image method for moving object detection. Besides, we propose a residual mechanism based on the raw point cloud to integrate the temporal and spatial information of the current scene.

\begin{figure*}[t]
\centering
\vskip 0.2in

\subfloat{
\includegraphics[width=7in]{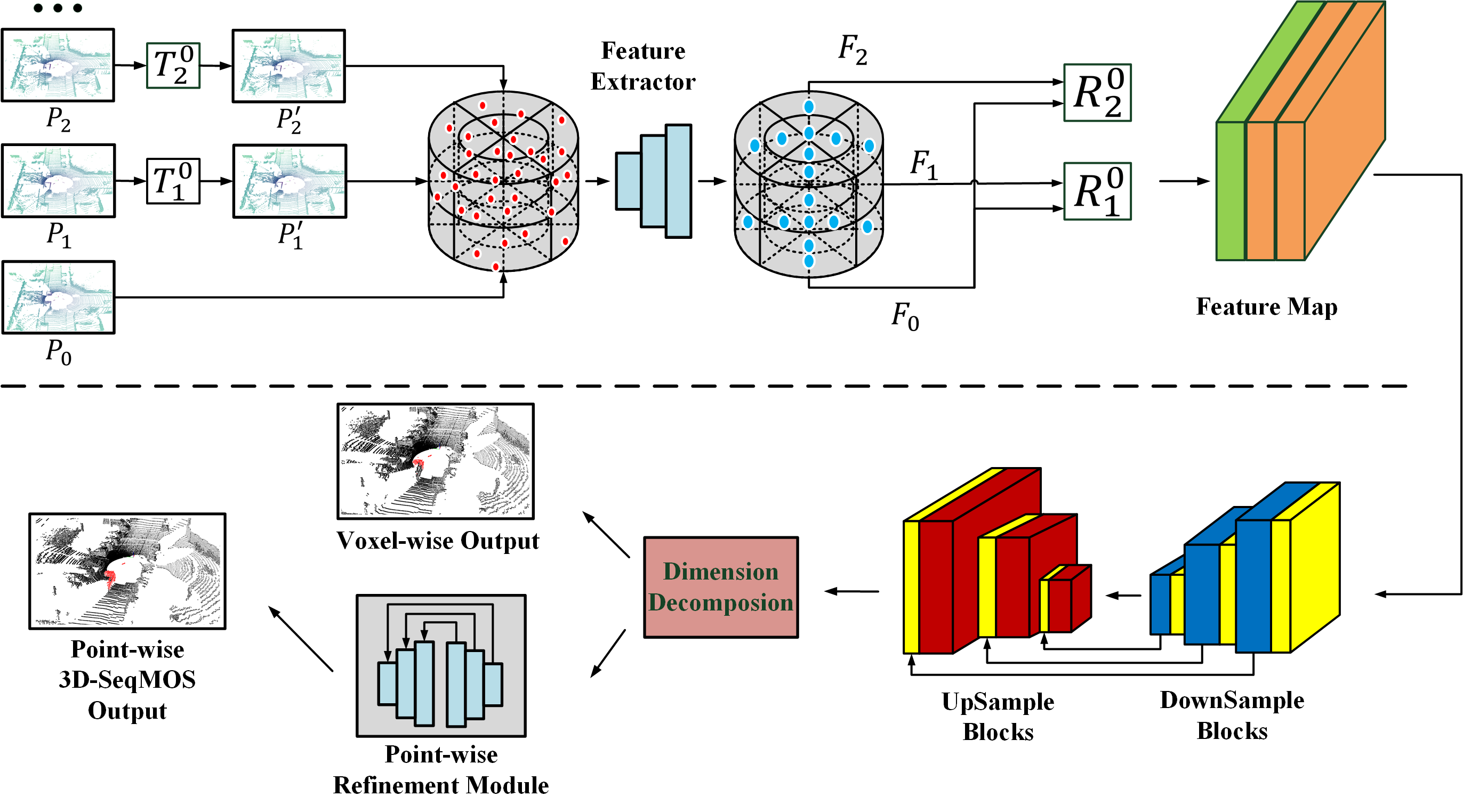}
\label{fig2.2}
}
\caption{Framework Overview of 3D-SeqMOS. Our network is divided into two parts: sequential point cloud feature extraction and moving point cloud segmentation, which correspond to the upper and lower parts of the figure respectively. In the feature extraction, we add the residual transformation of the raw 3D point cloud (red points) and use cylindrical partition to obtain the scan feature (block points). The feature map is combined current frame feature (green block) and the previous frames residual features (orange blocks). In the moving point cloud segmentation, both up sampling and down sampling contain three sub-modules. Each sub-module is composed of asymmetric blocks (yellow blocks) and convolution layers (red or blue blocks). Then, the segmentation label of each cylindrical voxel is obtained through the dimension decomposition module. In order to calculate the point-wise segmentation label, a refinement module is introduced to obtain the final output, that is, the motion label of each point, such as moving or non-moving.}
\label{fig2}
\end{figure*}

\subsection{LiDAR-SLAM}

SLAM is the basis of intelligent robot and autonomous driving. Especially in complex scenes such as dark light or lack of texture, LiDAR-SLAM has better robustness and stability than Visual-SLAM \cite{huai2021consistent, chou2021efficient}. And LiDAR-SLAM can explore the unknown environment without any prior information, unlike other sensors which need to deploy base stations in advance \cite{cao2021universal,xu2021distributed}. The accuracy of the front-end odometry and back-end loop-closure detection determines the performance of the whole autonomous driving system. Loam \cite{zhang2014loam} was a advanced SLAM system based on LiDAR. On this basis, a variety of improvement methods were proposed \cite{shan2018lego, chen2020sloam, zhang2015visual, wu2020robust}. To estimate self-motion, loam \cite{zhang2014loam}, lego-loam \cite{shan2018lego} extracted corner and plane features from the raw point cloud, and then minimized the distance from point to line and point to plane. However, due to the lack of practical loop-closure module, most algorithms cannot eliminate the cumulative error and built a globally consistent mapping. Lego-loam [18] segmented the raw point cloud, filtered out small clusters, and estimated the attitude through two-step Levenberg Marquardt optimization method. Although it can filter out some dynamic object, many static objects were reluctantly deleted, resulting in the loss of useful features. Moreover, they operated at the low-level feature data, which was unsuitable for reliably executing high-level instructions, such as semantics. Recently, some researchers have combined semantic information to assist point cloud registration. Zaganidis \emph{et al}. \cite{zaganidis2018integrating} extended normal distribution transformation (NDT)  \cite{biber2003normal} and generalized iterative closest point (GICP) \cite{segal2009generalized} using semantic information. Suma \cite{bai2019simgnn} was a surf based method with circular closure. Suma++ \cite{chen2019suma++} extended it with semantics to filter out dynamic object, so as to achieve the most advanced performance. Fast-Lio \cite{xu2021fast} keeped the raw point cloud data in dynamic balance by using the iKD-tree \cite{cai2021ikd} data structure, so as to use the raw points for odometry registration without extracting features.

Our work does not extract the point cloud features of planes and corners. Inspired by Fast-Lio2 \cite{xu2022fast}, we register the raw points to the local submap and then update odometry position and orientation without extracting features, which can be naturally applicable to different types of LiDAR. However, the Fast-Lio2 has no loop-closure detection module, so we add the scan-context descriptor to optimize the global pose. Combined with our 3D-SeqMOS, we reconstruct the whole end-to-end SLAM framework.

\section{Methodology}
In this paper, we propose a novel convolutional neural network framework (3D-SeqMOS) for 3D point cloud moving object segmentation, and build an improved SLAM system, including front-end odometry and back-end loop-closure detection. Next, we describe the details of our proposed method.


\subsection{Framework Overview of 3D-SeqMOS}
Our goal is to achieve accurate and effective moving object segmentation in LiDAR scanning, and reduce the impact of moving object on mapping and positioning. To preserve the structure information and context topology information of the 3D scans, we use the raw point cloud for moving object segmentation, rather than converting the point cloud into 2D image for convolution neural network processing. As shown in Fig. \ref{fig2}, our framework consists of two major components, including cylindrical feature extraction of spatio-temporal point cloud information and 3D convolution network for semantic segmentation of moving point cloud, corresponding to the upper and lower parts of Fig. \ref{fig2} respectively. Firstly, we use the residual mechanism to transform the previous scans to the local coordinate of the current scan, and take the multi-scans temporal residual information and the current scan spatial information as the input of the whole network. Then the input point clouds are divided by the cylindrical partition and the features extracted by multi-layer perceptrons (MLP) is reassigned based on the partition. Next, asymmetrical 3D convolution networks are used to generate the voxel-wise moving object segmentation outputs. However, for point cloud moving object segmentation task, we need each point segmentation outputs, so a point-wise refinement module is introduced to calculate the final moving object segmentation output labels.

\begin{figure*}[t]
\centering
\vskip 0.2in
\subfloat{
\includegraphics[width=7in]{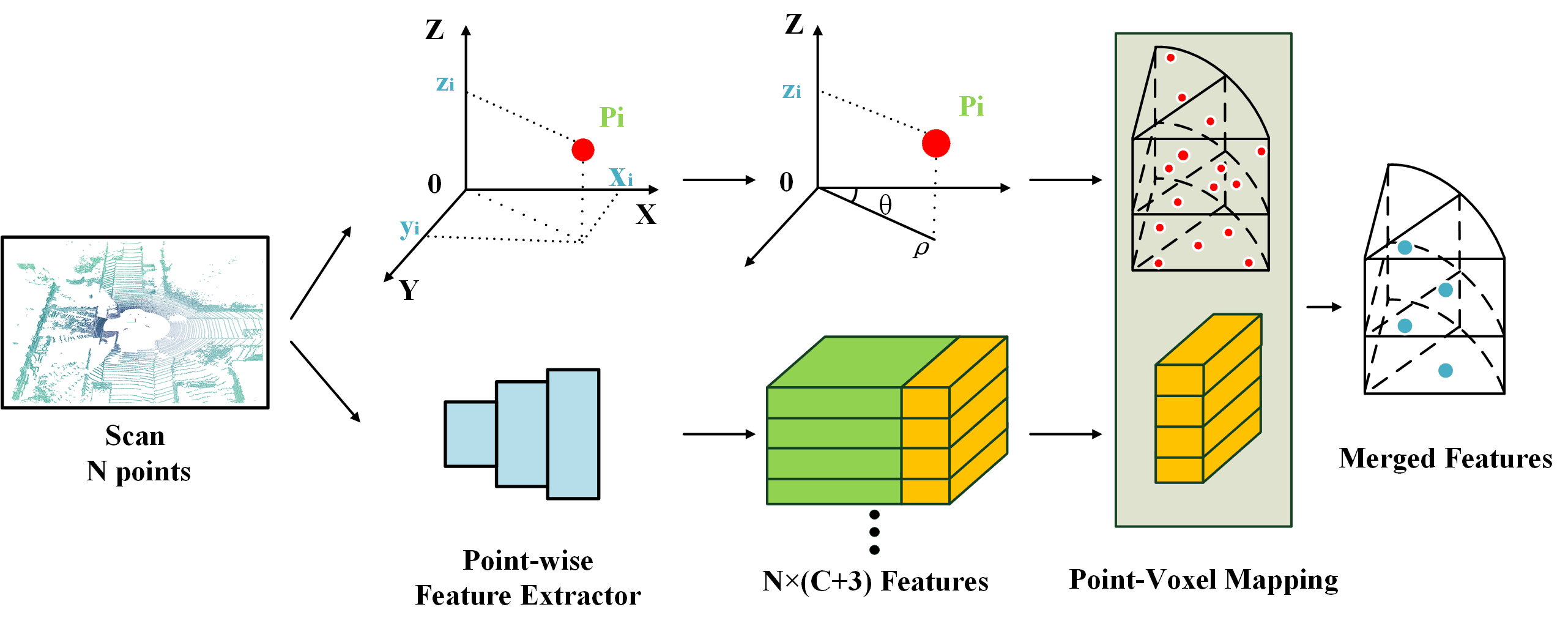}
\label{fig3}
}
\caption{Cylindrical Partition and Feature Extraction. It mainly includes two parts: cylindrical voxel partition and point feature extraction. Firstly, the raw N points are transformed into polar coordinate representation and then are mapped to a fixed size cylindrical voxel based on its polar coordinate (see the upper of the Point-Voxel Mapping). Next, MLP is used to extract the N+3 dimension features of raw N points, where C denotes the point feature dimension (green block) and 3 denotes the index of the current point in the cylindrical voxel (orange block). The mapping relationship between points and voxels can be obtained through the partition index. Finally, all features of each voxel are merged into a global feature (blue point).}
\label{fig3}
\end{figure*}

\subsection{Temporal Information and Residual Point Clouds}
Inspired by Wang \emph{et al}. \cite{wang2018temporal} and LiDAR-MOS \cite{chen2021moving}, which use the difference between RGB video frames for motion recognition, we propose to use the residual scan based on LiDAR to segment moving objects. Combined the current scan spatial information and the previous residual point cloud temporal information, our 3D-SeqMOS network distinguishes the semantic labels of moving objects and backgrounds.

We aim at segmenting moving objects, \emph{i.e.}, only using the current and recent LiDAR scans, such that one can exploit the information for odometry estimation in a SLAM pipeline and potentially remove moving objects from a map representation. We assume that we are given a time series of \emph{N} LiDAR scans in the SLAM history, denoted by \emph{P$_i$} = $\left\{p_j \in R^4   \right\}$, \emph{i = 1,2,...,N j = 1,2,...,M}, with \emph{M} points represented as homogeneous coordinates. We denote the current LiDAR scan by \emph{P$_0$} and the previous sequence of \emph{N} scans by \emph{P$_i$} with 1 \textless \, \emph{i} \textless \, \emph{N}. The estimated \emph{N} consecutive relative transformations from the SLAM odometry approach, $T_{N-1}^{N}  , . . . , T_1^0$, between the \emph{N+1} scanning poses, represented as transformation matrices, $T_k^h \in R^{4 \times 4}$ are also available.Given the estimated relative poses between consecutive scans, we can transform points from one viewpoint to another. We denote the \emph{m-th} scan transformed into the \emph{n-th} scan’s coordinate frame by
\begin{equation}
    P^{m\rightarrow n} = \left\{T_m^n p_i |p_i \in P_m  \right\}
\label{eq1}
\end{equation}
where $T_m^n = \prod_{k=m}^{n+1} T_k^{k-1}$.

To generate the previous scans and then fuse them into the current point cloud coordinate system, it needs to be transformed and re-projected. To realize this, we compensate for the ego-motion by transforming the previous scans into the current local coordinate system given the transformation estimates as defined in Eq.(\ref{eq1}). Therefore, as shown in Fig. \ref{fig2}, we convert \emph{P$_1$} and \emph{P$_2$} to \emph{P$_0$} coordinate system to obtain \emph{P$_1^{'}$} and \emph{P$_2^{'}$} respectively. Next, we use cylindrical partition segmentation and MLP to extract the features of the corresponding point cloud scans. And the \emph{R$_i^0$} denotes the residual feature of the previous \emph{i-th} scan relative to the current scan.

\subsection{Cylindrical Partition and Feature Extraction}
As we all know, the point cloud scanned by rotating machinery LiDAR in the outdoor scene has the characteristics of different density, that is, the density of the nearby area is much greater than that of the distant area. Therefore, we use cylindrical coordinate system to replace Cartesian mesh generation. It uses a grid that increases with distance to cover more areas, so as to more evenly distribute the points in different areas and match the distribution of outdoor points. As shown in Fig. \ref{fig3},  firstly, we convert the points in Cartesian coordinate system into cylindrical coordinate system, and calculate the radius in cylindrical coordinate system. This step converts points ($x_i, y_i, z_i$) to points ($\rho, \theta, z_i$). For the point cloud in each voxel, we use a multi-layer perception model to extract the each point feature to obtain the cylindrical features. Specifically, the point-cylinder mapping contains the index of point-wise features to cylinder. Based on this mapping function, point-wise features within same cylinder are mapped together and processed via max-pooling to get the cylindrical features, which is similar to the operation in PointNet \cite{qi2017pointnet++}. After these steps, we get the 3D cylindrical representation \emph{F $\in$ C $\times$ H $\times$ W $\times$ L,} where \emph{C} denotes the feature dimension and \emph{H, W, L} denote the radius, azimuth, and height, respectively.

\begin{figure}[t]
\centering
\vskip 0.2in
\subfloat{
\includegraphics[width=3.4in]{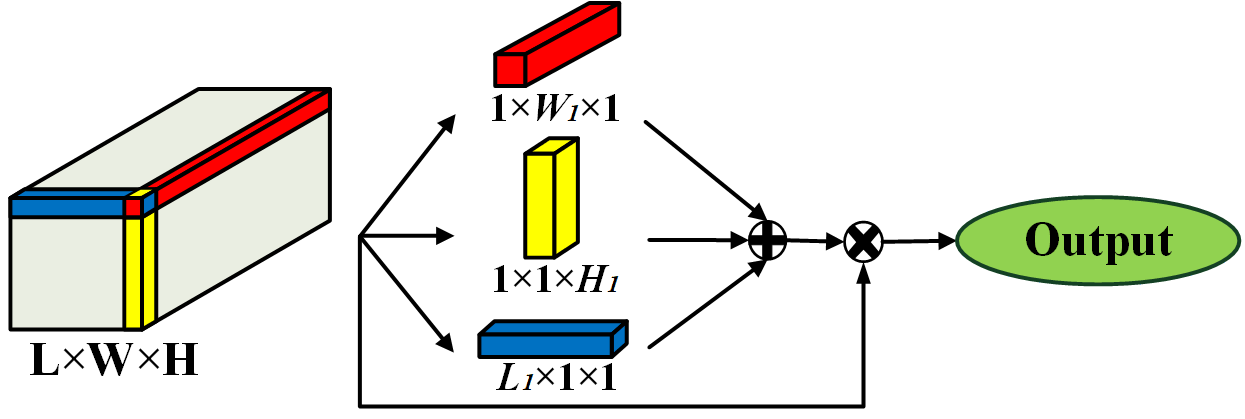}
\label{figDDCM}
}
\caption{Dimension-Decomposition Module. For the \emph{L} $\times$ \emph{W} $\times$ \emph{H} input sensor, we use three dimension decomposition convolution kernels to extract feature information of different dimensions, and aggregate the extracted features with the input to obtain the final global context information.}
\label{fig.DDCM}
\end{figure}

\subsection{Asymmetrical Sample and Dimension-Decomposition}
The driving scene point cloud performs specific object shape distribution. In this paper, it mainly includes other cube objects such as moving cars, trucks, buses, motorcycles and pedestrians, so our goal is to follow this observation to enhance the representation ability of standard 3D convolution. In addition, recent literature \cite{ding2019acnet} and Cylinder-3D \cite{zhu2021cylindrical} also show that the central crisscross weight plays a greater role in the square convolution kernel. Therefore, we use asymmetric residual blocks to strengthen the horizontal and vertical response, and match the distribution of target points, and use asymmetric up-sampling blocks and asymmetric sample loading blocks for sample unloading and loading operations. As shown in Fig. \ref{fig2}, the up-sampling and down-sampling both contain three sub-modules, each of which is composed of an asymmetric residual block and a convolution layer. On this basis, a dimension-decomposition is introduced to explore the high-order global context in the decomposition aggregation strategy, as shown in Fig. \ref{fig.DDCM}. We follow the tensor decomposition theory \cite{chen2020tensor} and cylinder voxel decomposition to build a high-rank context as a combination of low-rank tensors, where we use three rank-1 kernels in the three dimensions of feature space to obtain low-rank features, and then aggregate them together to obtain the final global context, as shown in Fig. \ref{fig.DDCM}.

\subsection{Point-wise Refinement Module}
The voxel-based method predicts a label for each cell. Although this method effectively explores a wide range of point clouds, it is inevitably affected by lossy cell label coding. There are many different labels of points in the same cell, which will lead to the loss of information of point cloud semantic segmentation task. So it is necessary to introduce point-wise refinement module to reduce the interference of lossy cell label encoding and refine the semantic label of each point. We process the cylindrical-voxel features to the point-wise based on the inverse Point-Voxel Mapping. And we project that the points of same voxel partition would be regarded as the same voxel feature. Then we integrate both point features before and dimension decomposition output as the input of the point-wise refinement network and fuse them together to refine the output as the final 3D-SeqMOS output.

\begin{figure}[h]
\centering
\vskip 0.2in
\subfloat{
\includegraphics[width=3.4in]{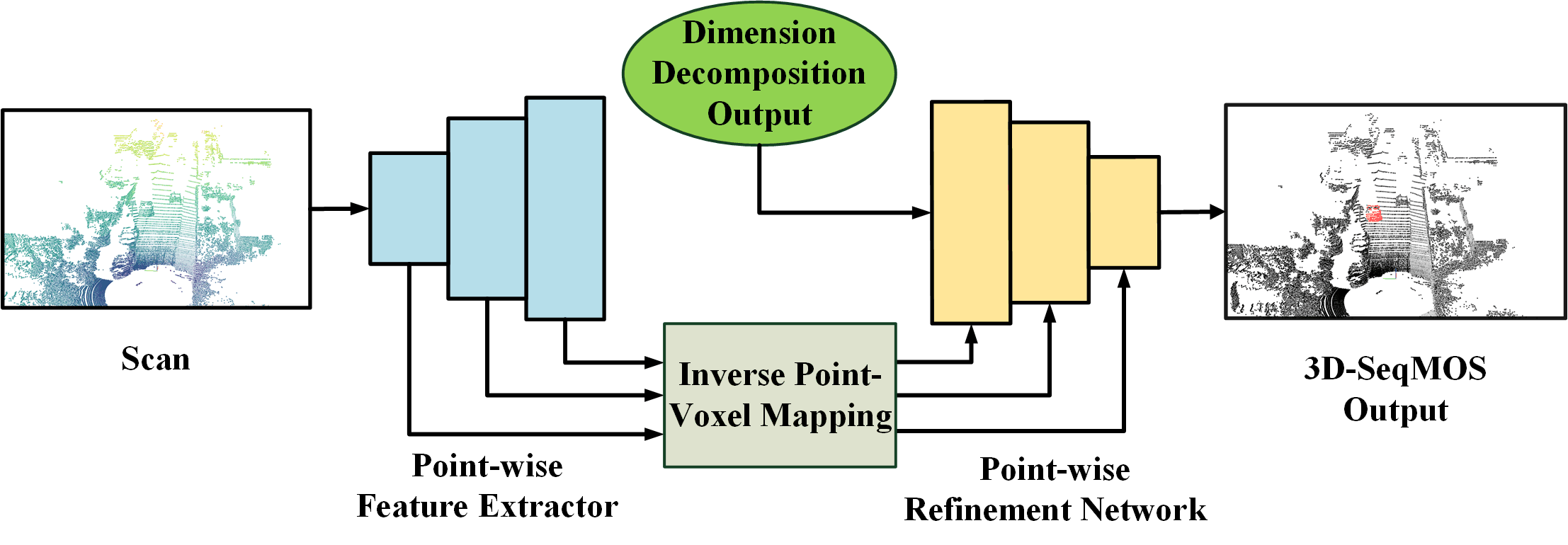}
\label{figDDCM}
}
\caption{Point Refine Module. Before we use the point-wise feature extractor (blue blocks) to get the cylinder voxel features. According to inverse Point-Voxel-Mapping, we project the cylinder voxel features to each point. Then, both the dimension decomposition output (green block) and the point-wise feature are taken as the input of point-wise refinement network (yellow blocks), and then are fused together to optimize the points label output, as the final whole 3D-SeqMOS.}
\label{fig.refine}
\end{figure}

\subsection{Loss Function}
The loss for the moving semantic object segmentation task based on LiDAR includes two parts: point-wise loss and voxel-wise loss. The total loss of our network can be defined as:
\begin{equation}
    L_{total} = \alpha L_{point} + \beta L_{voxel}
\label{loss}
\end{equation}
where $\alpha$  and $\beta$ represent the weight of corresponding loss, respectively. In our paper, we use the same weights for training. For voxel-wise loss $L_{voxel}$, we follow the existing methods \cite{hu2020randla} and  \cite{cortinhal2020salsanext}, and use the weighted cross-entropy loss and lovasz-softmax \cite{berman2018lovasz} loss to maximize the point accuracy. For point-wise loss $L_{point}$, we only use weighted cross entropy loss to supervise training. In the forward inference process of our model, the output of the point by point refinement module will be used as the final output of our 3D-SeqMOS. For the optimizer, we use the Adam with an initial learning rate of 0.001. Besides, for each verification, we all trained more than 30 epochs.

\subsection{SLAM Frame}
At present, the mainstream LiDAR odometry measurement method is mainly based on Loam \cite{zhang2014loam}. To improve the operation efficiency, these methods extract the plane features and corner features of each scan for matching, which will lose a lot of useful information of 3D point cloud. Inspired by the iKD-tree \cite{cai2021ikd} and Fast-Lio \cite{xu2021fast}, we can no longer extract the above plane and corner features at the front-end of the odometry, but directly register the raw 3D point cloud. In the front-end odometry, we adopt the based on Fast-Lio and combine it with our proposed 3D-SeqMOS. Through the direct registration of the raw point cloud by the iKD-tree data structure, our 3D-SeqMOS can filter out the moving objects point cloud in the autonomous driving scene, which can not only alleviate the amount of calculation, but also reduce the interference to the point cloud registration. Through the above operations, we can get the preliminary odometry pose estimation results. However, the based on Fast-Lio odometry method does not support the loop-closure, so we propose the back-end loop detection module for our 3D-SeqMOS and odometry. To verify the robustness of the overall SLAM system and the effectiveness of our proposed 3D-SeqMOS, we design a loop-closure detection module, which is based on scan context geometric descriptor. The overall framework is shown in the Fig. \ref{fig4}.

\begin{figure}[h]
\centering
\vskip 0.2in
\subfloat{
\includegraphics[width=3.4in]{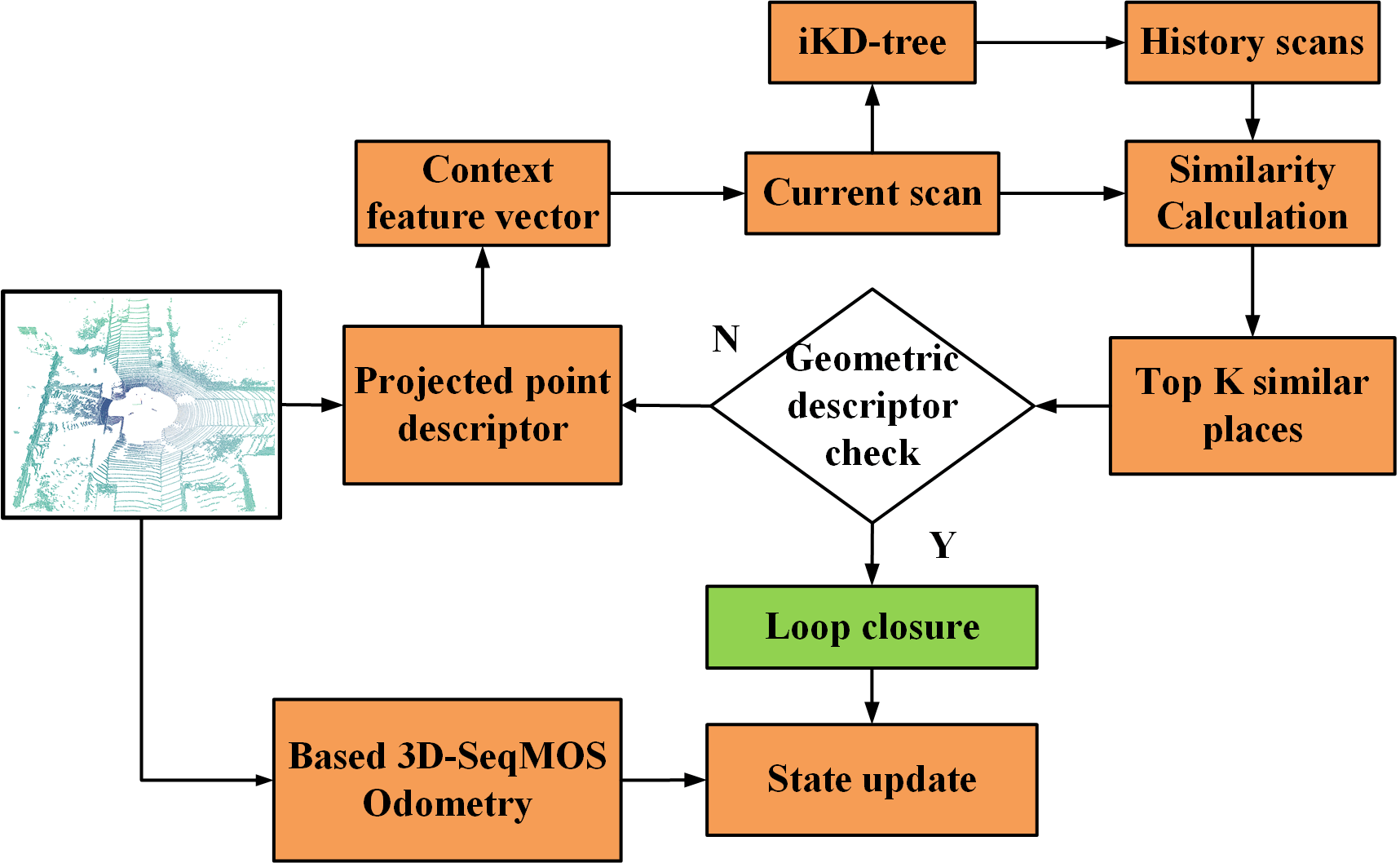}
\label{fig4}
}
\caption{The SLAM frame of back-end loop-closure detection and pose optimization.}
\label{fig4}
\end{figure}

In order to improve accuracy of the proposed system, based on Fast-Lio odometry, we add a loop-closure detection module. Firstly, for current scan, the preliminary system position can be calculated through the odometry module. Then, in loop-closure optimization, we design a feature descriptor vector for the current scan. The horizontal axis of the vector represents the angle value of the laser scanning, and the vertical axis represents the depth ring value. The pixel value is the average height of all points in the intersection area of the corresponding angle and the number of depth ring. By transforming current 3D point cloud data into a 2.5D context structure, the data dimension and the amount of calculation are reduce. And the amount of data is reduced by about 1000 times without losing the core information of the point cloud.

In addition, to quickly match the current scan with the historical key frames, we use the iKD-tree structure for data search and maintenance. Then calculate the similarity between the current frame and the historical key frame, and select the top k key frame positions as the candidate for closed-loop detection. Next, by setting the time threshold and geometric feature descriptor to determine whether to carry out closed-loop detection, and finally update the optimize the global state. In test experiments, we integrate our 3D-SeqMOS and LiDAR-SLAM module into a complete end-to-end framework. It is noted that, our 3D-SeqMOS can be perfectly combined with the existing SLAM method.

\section{Experiments}
In this section, we first provide detailed experiments and data settings, and then evaluate the proposed 3D-SegMOS method on the large-scale outdoor scene. Our paper mainly focuses on 3D moving object segmentation from raw LiDAR scan sequences. We will conduct a  number of experiments to show the performance of our method and the improvement for SLAM positioning and mapping. We use the odometry information provided by SemanticKITTI, which is estimated by the LiDAR based SLAM system Suma \cite{chen2019suma++}. SemanticKITTI is a large-scale driving scene data set for point cloud segmentation. We divided 22 sequence data sets: sequences 00 to 10 were divided into training sets (in which sequence 08 is used as verification set), and sequences 11 to 21 were divided into test sets. In the raw point cloud, each point has its corresponding semantic label with total 34 categories in dataset. Here, we are more interested in separating points into moving and static categories.

\begin{figure}[b]
\centering
\vskip 0.2in
\subfloat{
\includegraphics[width=3.4in]{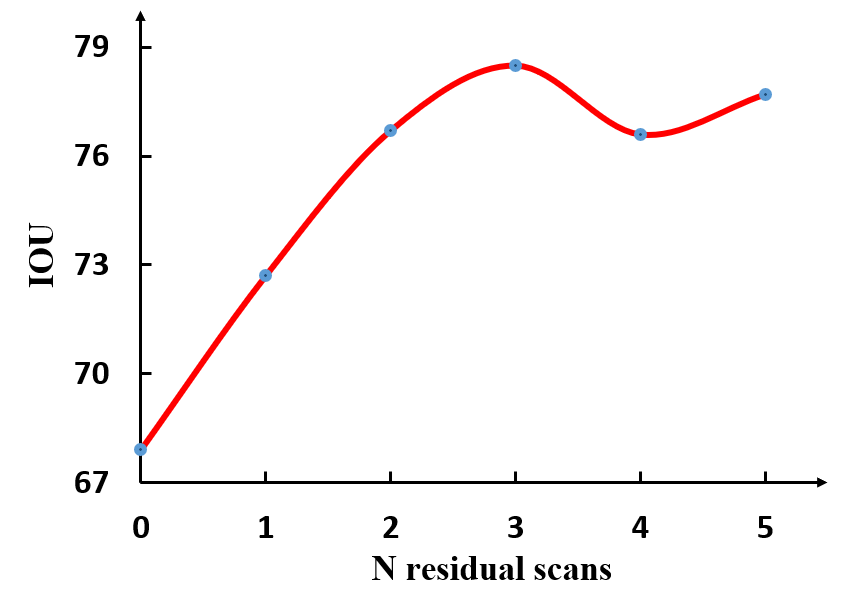}
\label{fig5}
}
\caption{Ablation studies. This figure shows the relationship between 3D-SeqMOS IOU and the number of residual scans.}
\label{fig5}
\end{figure}

\begin{figure*}[t]
\centering
\subfloat[]{\includegraphics[scale=0.23]{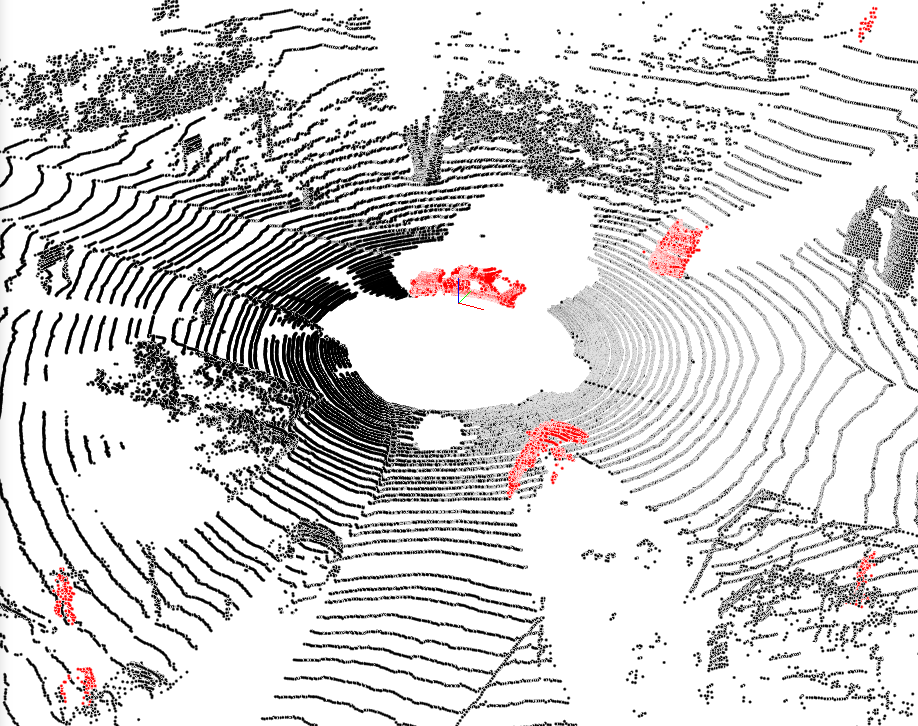}%
\label{fig6.1}}
\hfil
\subfloat[]{\includegraphics[scale=0.23]{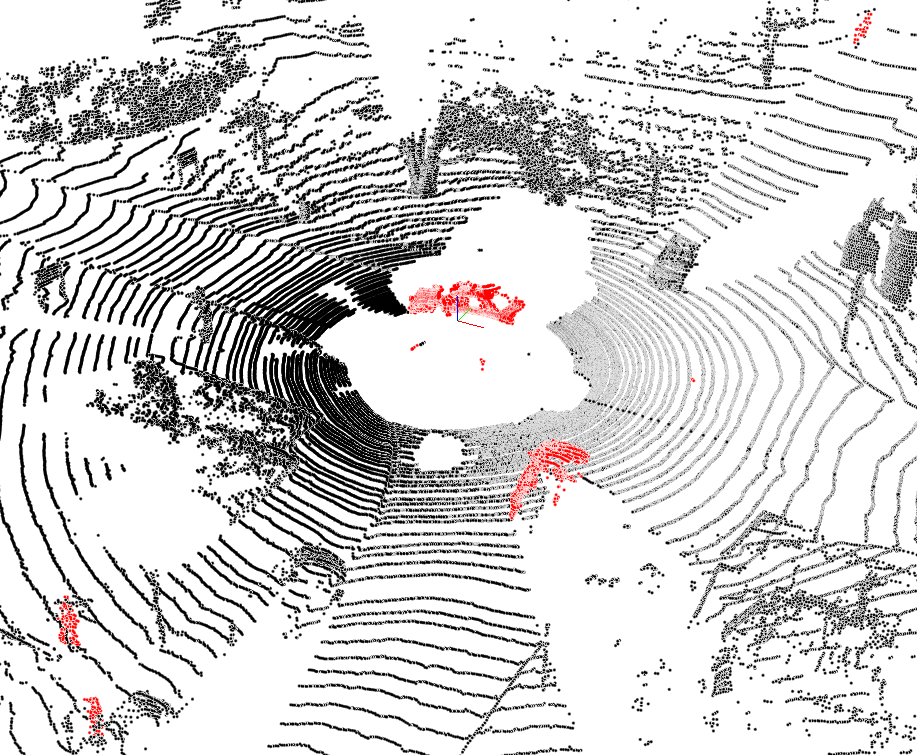}%
\label{fig6.2}}
\hfil
\subfloat[]{\includegraphics[scale=0.23]{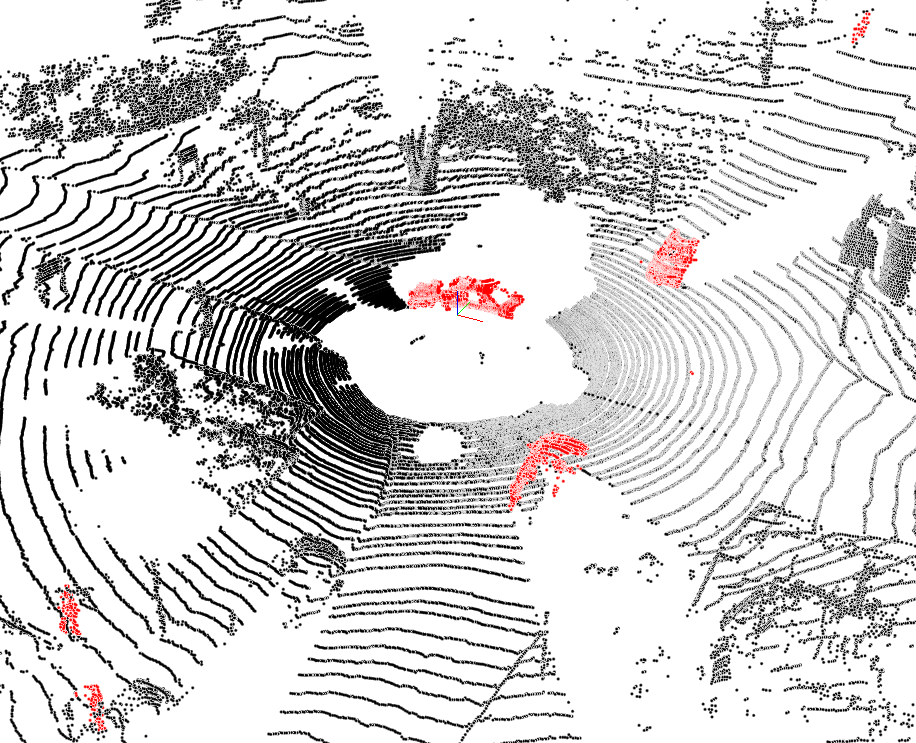}%
\label{fig6.3}}
\hfil
\subfloat[]{\includegraphics[scale=0.23]{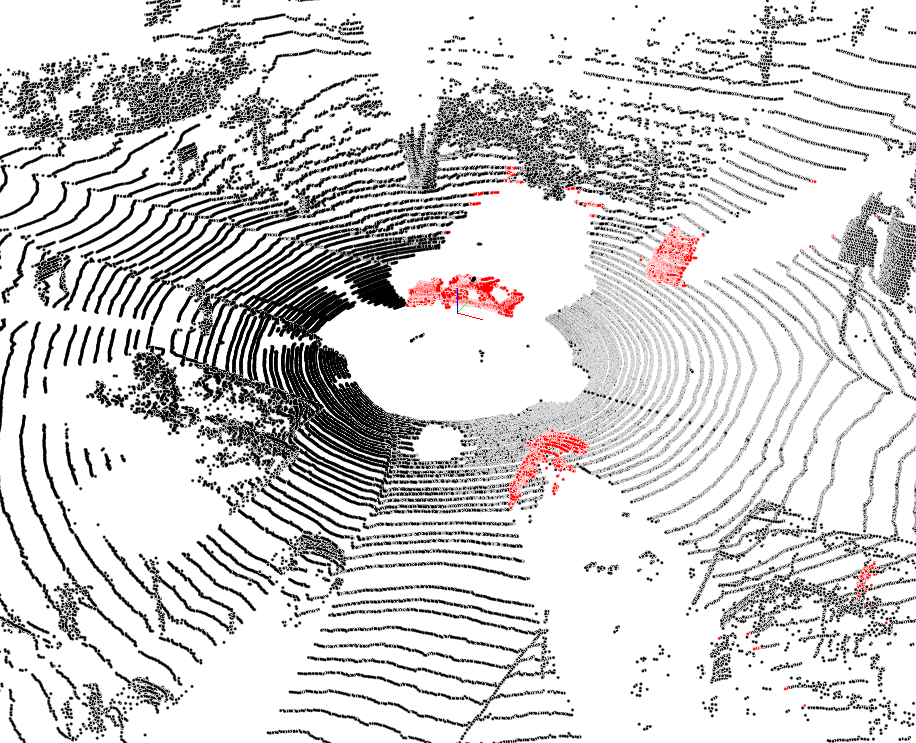}%
\label{fig6.4}}

\caption{The results of the moving object segmentation performance on SemanticKITTI dataset, the moving object is colored red. (a) Ground truth, (b) our result (no residual), (c) our result (3 residual) and (d) result of the LiDAR-MOS.}
\label{fig_6}
\end{figure*}

\subsection{Ablation Study on Input}

The ablation study introduced in this section aims to support our point cloud residual view that our method can segment moving objects only using 3D LiDAR scans. We tested on validation dataset sequence 08. We compared experiments, firstly, we only used the model trained with the current one scan, and next, we setup the current point cloud scan and three previous frame residual point cloud as the input to calculate the results of the training model. When fusing residual information, because the number of points of point cloud is different between two frames, we can choose to fuse directly, and then make residual in the feature layer. As shown in the Fig. \ref{fig6.2} and Fig. \ref{fig6.3}, when only the current single scan is used as the input, the segmentation results are worse, such as incomplete segmentation of some vehicles and wrong recognition. The structure with residuals is obviously better than the single scan result, so next, we conduct ablation experiments to get the residual frames corresponding to the best results.

We still follow the previous dataset settings and test on verification dataset sequence 08 for selecting the better number of residual frame. In each experiment, the residual frames of the previous point cloud are gradually added, and then calculate final moving label mean intersection-over-union value. Almost every training experiment has exceeded 30 epochs to ensure the stability of the experimental results. As shown in the Fig. \ref{fig5}, by ablation experiments on different residual frames and considering the size of data, we believe that our 3D-SeqMOS can achieve the best results by fusing 3 frames residual. 

\subsection{Moving Object Segmentation Performance}

This section analyzes the moving object segmentation performance of the proposed method through experiments. We compare with some mainstream moving object segmentation algorithms. We use 00-10 sequences of SemanticKITTI for training, except 08 for verification, and used 11-21 to test by uploading our results to \emph{SemanticKITTI: Moving Object Segmentation}\footnote{https://competitions.codalab.org/competitions/28894} ranking $2nd$ on the leaderboard. It should be noted that, according to our ablation experiment, we use the current frame scan and the previous 3 frame residuals as input to train our model. Since there are few existing implementation methods of LiDAR based MOS, according to  \cite{chen2021moving}, we select several methods that have been used in similar tasks, such as semantic segmentation and scene flow, and modify them to implement moving object segmentation. All methods are based on the proposed benchmark data, that is, the evaluation of test sequence 11-21. In particular, these algorithms all process moving object segmentation by converting 3D point clouds to 2D images. Our proposed method may be the first to use the raw 3D point cloud for moving point cloud segmentation. 

To qualitatively compare and analyze the effectiveness of our method, we visually compare our results with the results of the current more advanced method LiDAR-MOS. We compare the effect of moving object segmentation on the same scene. We compare with the current more advanced LiDAR-MOS method. To verify the generalization ability of our method to the autonomous driving scene, we choose different scenes in the verification set and the test set, including campus and highway. Firstly, we make a comparison on the complex traffic crossroad of verification-set sequence 08 with ground truth labels, including a large number of moving cars, waiting static cars and small pedestrians. It can be seen from the Fig. ~\ref{fig_6} that our method has an excellent effect on the recognition accuracy and target integrity. For example, the cars recognized by LiDAR-MOS are incomplete, some small and distant moving object are not recognized. Beside, both methods can correctly identify the waiting static car.

It can be seen that in a challenging actual scene, there are a large number of moving and stationary object point clouds at the same time. The LiDAR-MOS result based on 2D images has large errors, such as incomplete object contour, inaccurate recognition of distant object, and even wrong recognition of some static point clouds as moving point clouds. In comparison, our method can well identify the moving object point cloud, but only the pedestrians far away in the lower right corner of the picture cannot be recognized. Our method can distinguish between moving points and static points, even if some moving objects move slowly, which cannot be detected by other methods.

To evaluate the proposed method, we follow the official guidance to leverage mean intersection-over-union (\emph{mIoU}) as the evaluation metric defined in \cite{landrieu2018large} formulated as:

\begin{equation}
 IoU_i = \frac{TP_i}{TP_i+FP_i+FN_i} 
\end{equation}

Where \emph{TP$_i$, FP$_i$, FN$_i$} represent true positive, false positive, and false negative predictions for class \emph{i} and the \emph{mIoU} is the mean value of \emph{IoU$_i$} over all classes.

\begin{table}[h]
\begin{center}
\caption{Results of moving object segmentation with analyzing a variety of related semantic segmentation methods, which are taken from \cite{chen2021moving}.}
\label{table:1}
\begin{tabular}{lc}
\hline\noalign{\smallskip}
Method & \qquad \qquad \qquad \emph{mIoU}\\
\noalign{\smallskip}
\hline
\noalign{\smallskip}
{\upshape SalsaNext(retrained)}    & \qquad \qquad \qquad 46.6  \\
{\upshape Residual + RG + Semantics} & \qquad \qquad \qquad 20.6 \\
{\upshape SceneFlow + Semantics}    & \qquad \qquad \qquad 28.7 \\
{\upshape SqSequence}    & \qquad \qquad \qquad 43.2      \\
{\upshape KPConv}     & \qquad \qquad \qquad 60.9  \\
{\upshape LiDAR-MOS(N=1)}       & \qquad \qquad \qquad 52.0 \\
{\upshape LiDAR-MOS(N=8) }      & \qquad \qquad \qquad 62.5   \\
\textbf{Ours(no residual)}   & \qquad \qquad \qquad \textbf{59.8}  \\
\textbf{Ours(3 residuals)}   & \qquad \qquad \qquad \textbf{74.9}  \\
\hline

\end{tabular}
\end{center}
\end{table}

Table~\ref{table:1} shows some existing advanced moving object segmentation networks, such as Salsanext \cite{cortinhal2020salsanext}, KPConv \cite{thomas2019kpconv}, here, we also show the results generated by the retrained Salsanext and the binary tag as Salsanext (retraining). Sceneflow estimates the translation flow vector of each LiDAR point given two consecutive scans as input to extract difference. LiDAR-MOS is the latest research achievement and has the advanced performance. It is also a moving object segmentation method by converting 3D point cloud into 2D image, and our method is still 12.4\emph{\%} better than it. By comparing the last two rows of the Table~\ref{table:1}, we find that adding temporal information using residual point cloud and combining the spatial information of the current scan for moving object segmentation, we can see that the performance has been improved by 15.1\emph{\%}.

\begin{figure}[h]
\begin{center}
\subfloat[]{
\includegraphics[scale=0.17]{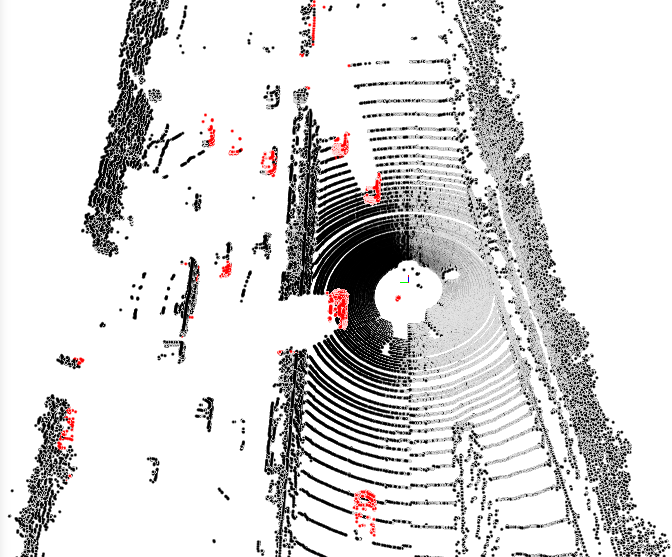}
\label{fig7.1}
}
\subfloat[]{
\includegraphics[scale=0.17]{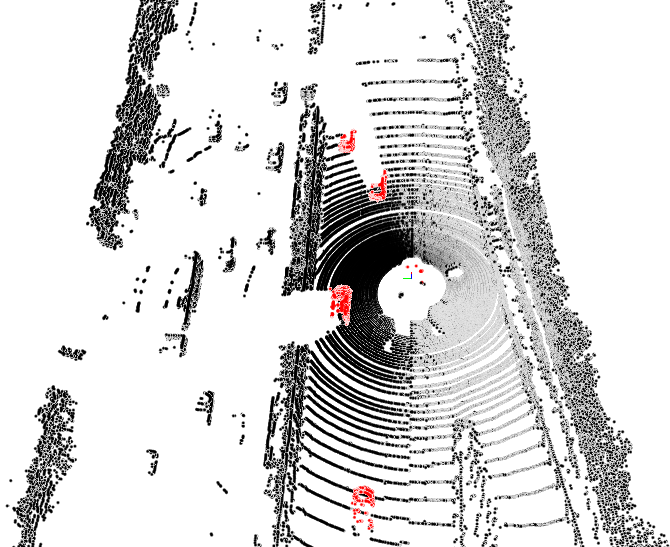}
\label{fig7.2}
}

\subfloat[]{
\includegraphics[scale=0.2]{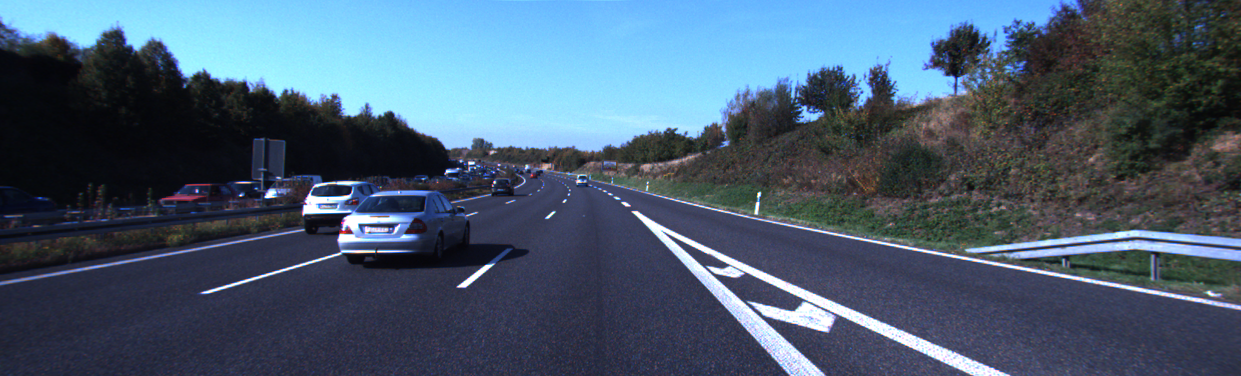}
\label{fig7.3}
}
\caption{The results of the moving object segmentation on  normal high-speed driving scene with the traffic jam, and the moving object is colored red. (a) is the LiDAR-MOS result, (b) is our result, (c) is colorful image.}
\label{fig7}
\end{center}
\end{figure}

Next, we analyze the generalization ability of our model in the highway scene. As we know, the stability and robustness of the highway scene has always been a more challenging problem in the autonomous driving. The high speed of vehicles and the large number when in traffic jam, both are difficult problems in autonomous driving. We will conduct comparative analysis on the testset sequence 21, and select the normal high-speed driving scene with the traffic jam. It should be noted that there is no truth label on the test set. Refer to SemanticKITTI to analyze the running state of the vehicle. The left road is full of vehicles blocked and stopped, and the right road is normal high-speed vehicles. As shown in Fig. \ref{fig7}, LiDAR-MOS has poor segmentation effect in the left complex road scene with more noise and mis-identification. Our method can accurately identify the 
moving vehicles in Fig. \ref{fig7.2}.





\begin{figure}[h]
\begin{center}
\subfloat[]{
\includegraphics[scale=0.58]{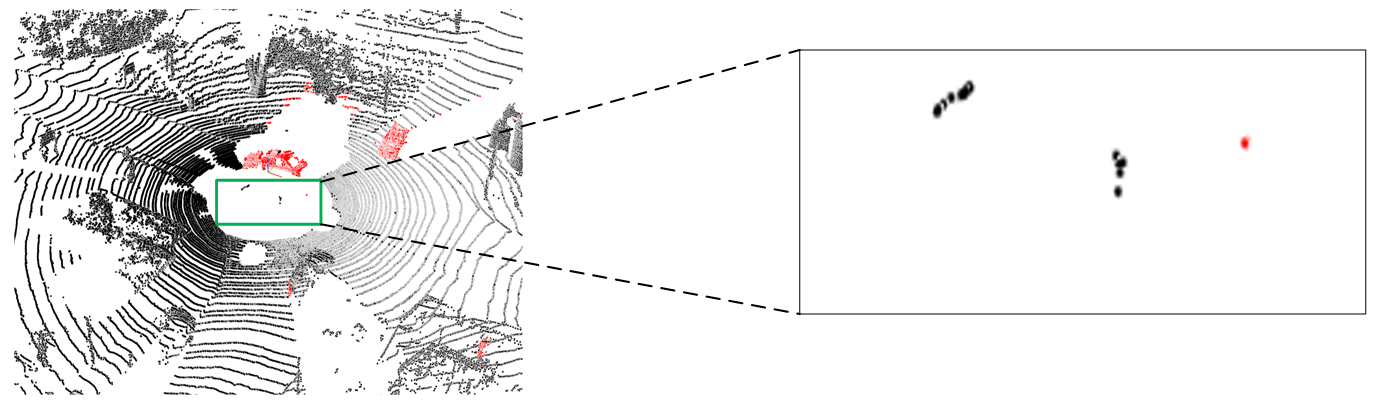}
\label{fig7.1}
}

\subfloat[]{
\includegraphics[scale=0.58]{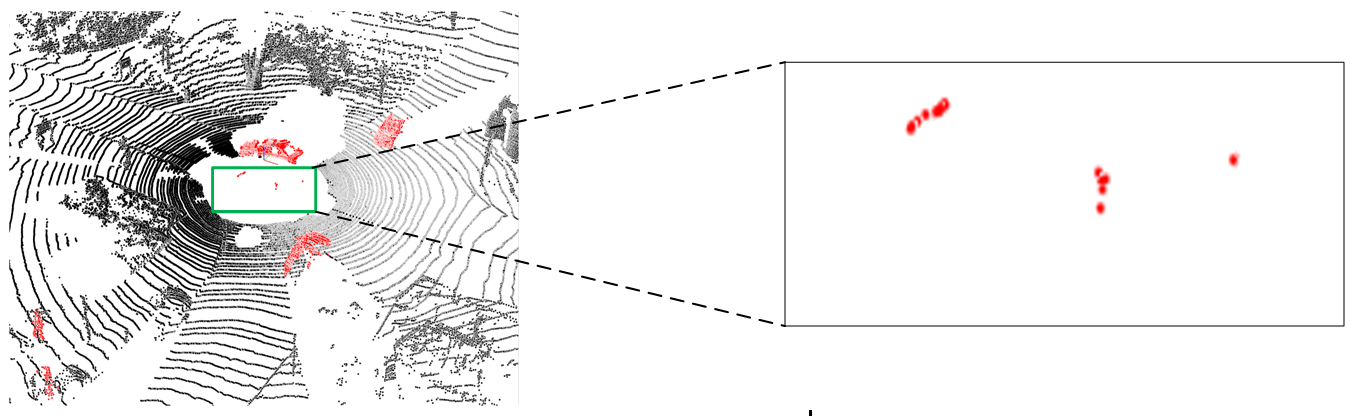}
\label{fig7.1}
}
\caption{The results of the moving object segmentation on acquisition vehicle, (a) is the LiDAR-MOS result, (b) is our result. The left road is full of vehicles blocked and stopped, and the right road is normal high-speed vehicles.}
\label{fig8}
\end{center}
\end{figure}

As shown in the Fig. \ref{fig8} we are also glad to find that our algorithm is very effective and sensitive to the segmentation of moving point clouds. In SemanticKITTI dataset, all of the point clouds are collected through Velodyne-64 which on the acquisition vehicle. The acquisition vehicle blocks the viewing angle of Velodyne-64 LiDAR, so a small amount of point clouds fall on the acquisition vehicle. Theoretically, these point clouds are also moving objects relative to the scene, and our method can also identify them, but LiDAR-MOS cannot. In ground truth, it is not considered as a moving point cloud. After analysis, we found that SemanticKITTI truth is a label assigned according to the object category, and these point clouds do not belong to any category, so they ignore these semantic label.

\subsection{Performance Improvement of LiDAR Odometry}
Our initial goal is to eliminate the influence of moving object in the scene on mapping and positioning, so we analyzed the improvement effect of our proposed method on LiDAR-SLAM positioning \cite{you2018joint}. As we all know that loop-closure and pose-optimization in SLAM plays a strong role in improving the trajectory, but in order to test the improvement effect of our method on positioning, we close the power of loop-closure. So if the performance can improve, it only comes from our proposed method. We selected two very representative methods for testing, Fast-lio2 \cite{xu2022fast}, which is the latest LiDAR odometry method, and Lego-loam \cite{shan2018lego}, which is widely used in the autonomous driving industry, and we carried out experimental analysis with our proposed segmentation method on the official SemanticKITTI dataset. From Fig. \ref{fig9} below, we can intuitively see the advantages of our method.

\begin{figure}[h]
\begin{center}
\subfloat[]{
\includegraphics[scale=0.65]{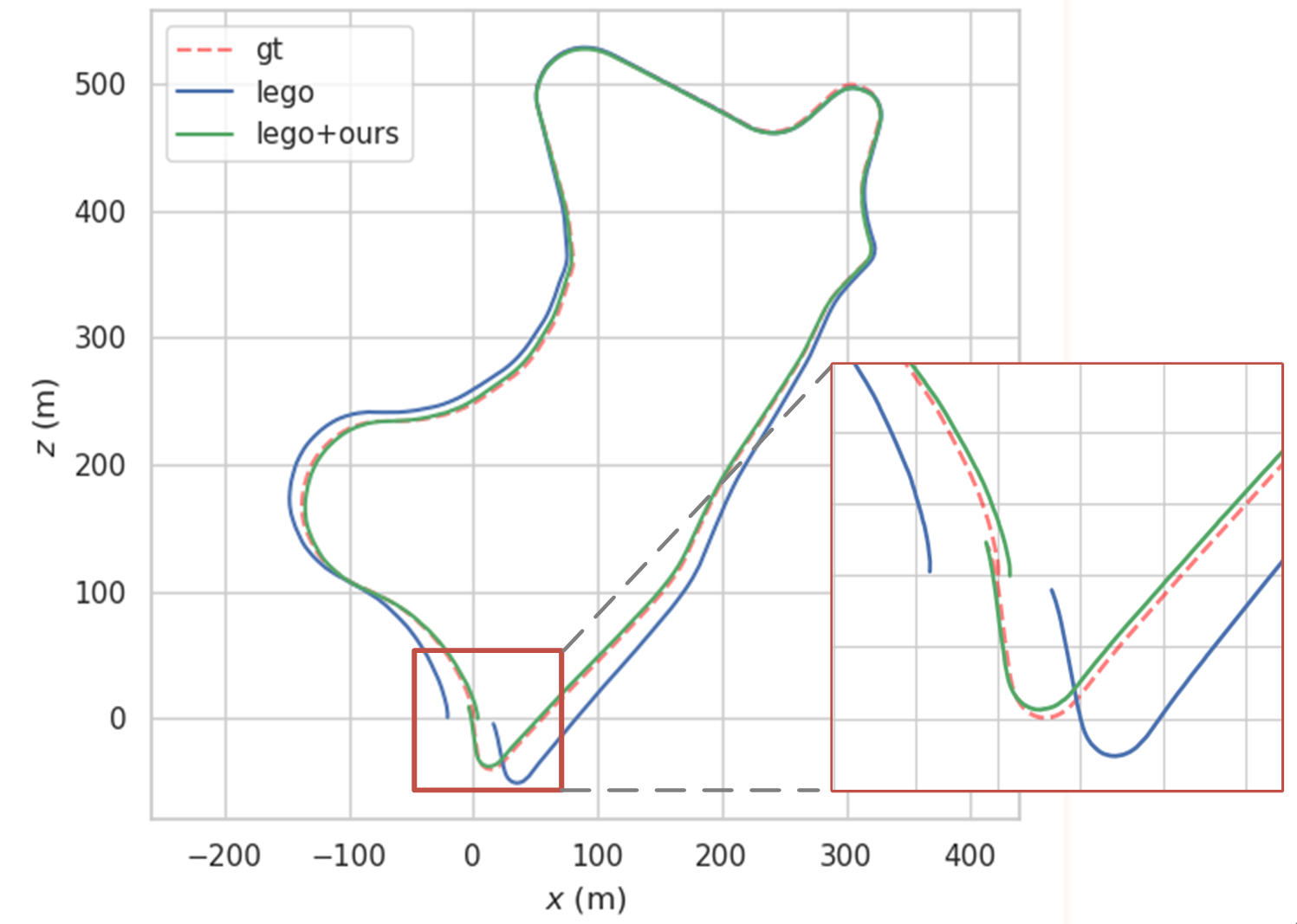}
\label{fig9.1}
}

\subfloat[]{
\includegraphics[scale=0.65]{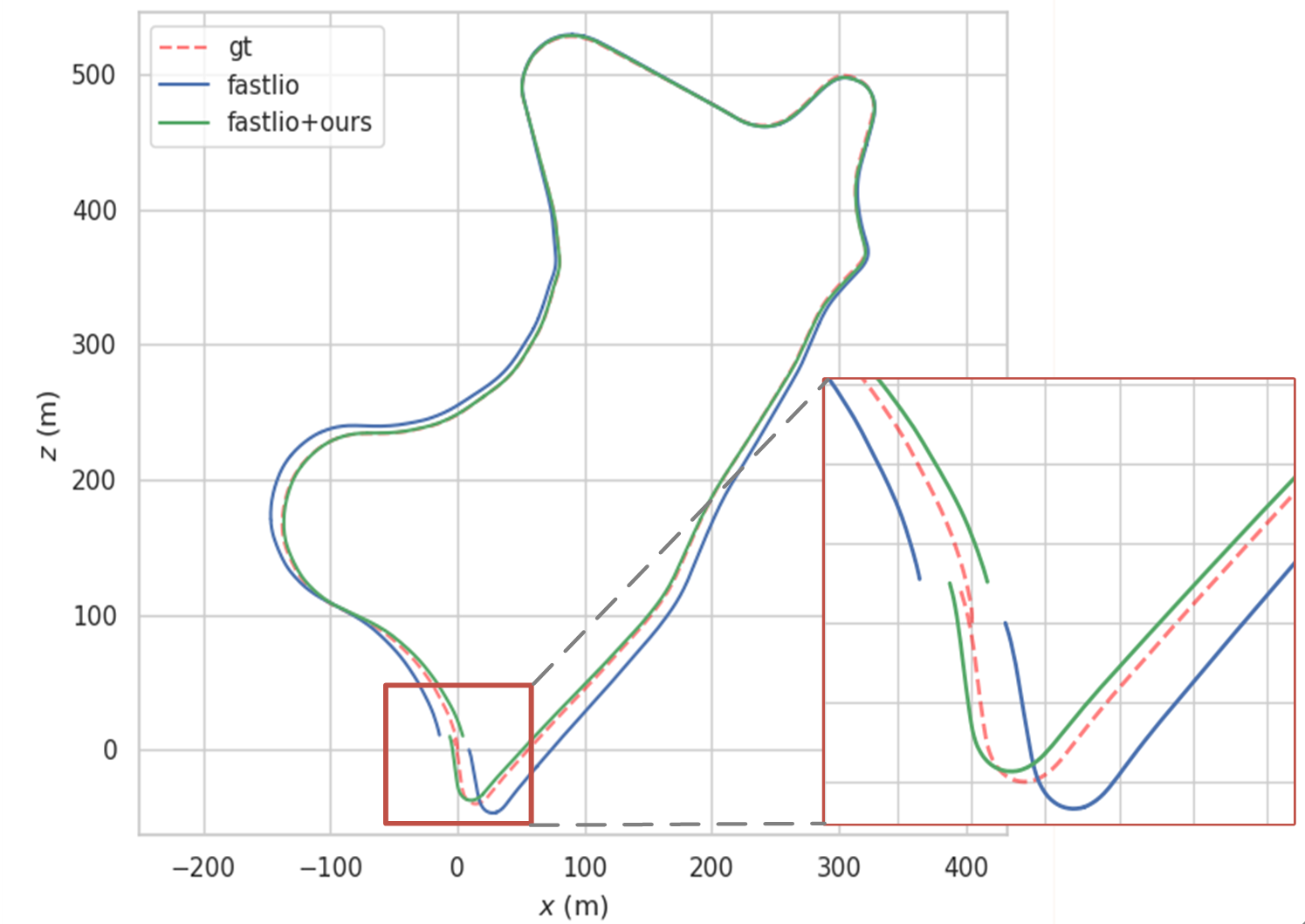}
\label{fig9.1}
}
\caption{The improved effect of filtering moving object for odometry. The red dotted line is the ground truth, the blue line is the result of the raw method, and the green line is the result of our optimization. It should be noted that for both methods, we have turned off the loop-closure function, and the improvement effect only comes from our proposed method. We can intuitively see that our proposed method can significantly improve the positioning accuracy.}
\label{fig9}
\end{center}
\end{figure}

\textbf{Evaluation position metric}. The direct evaluation criterion of the positioning algorithm is whether the algorithm can accurately estimate the driving track of the unmanned vehicle. The higher the global consistency between the estimated trajectory and the real trajectory, the more accurate the positioning algorithm is. The difference between the estimated pose and the real pose of continuous frames is usually measured by absolute trajectory error (\emph{ATE}), as follows:

\begin{equation}
  ATE = {\Big( \frac {1}{m}\sum _{i=1}^{m}({|trans(Q_i^{-1}P_i)|}^{2} \Big)} ^{\frac{1}{2}}
\end{equation}
where \emph{m} is the total scans number in the data, \emph{Q$_i$} $\in$ \emph{SE(3)}, \emph{i=1,2,…,m}, is the true pose of \emph{i-th} scan, \emph{P$_i$} $\in$ \emph{SE(3)}, \emph{i=1,2,…,m} is the predicted pose of \emph{i-th} scan, and \emph{trans()} represents the translation part in the absolute trajectory error. We believe that the absolute trajectory error can better reflect the positioning performance of different algorithms, and is more suitable for sensors with different sampling frequencies. In order to better illustrate our segmentation effect, we conduct quantitative analysis by positioning metric, see TABLE \ref{table:2}. We use absolute trajectory error \emph{ATE} metric, start-to-stop drift error (\emph{DE}) and drift rate (\emph{DR}) per meter for quantitative analysis. 

\begin{table}[t]
\begin{center}
\caption{The improved effect by our proposed method using \emph{ATE} error, the start-to-stop drift error (\emph{DE}) and average drift rate (\emph{DR}) of per meter.}
\label{table:2}
\begin{tabular}{llll}
\hline\noalign{\smallskip}
{\upshape Method} & \qquad \qquad \emph{ATE(m)} & \emph{DE(m)} & \emph{DR(\%)}\\
\noalign{\smallskip}
\hline
\noalign{\smallskip}
Lego & \qquad \qquad 15.37 & 36.33 & 2.11\\
Lego + \textbf{Ours} & \qquad \qquad \textbf{10.30} & \textbf{11.76} & \textbf{0.69}\\
Fast-lio2 & \qquad \qquad 7.15 & 24.29 & 1.42 \\
Fast-lio2 + \textbf{Ours} & \qquad \qquad \textbf{2.32} & \textbf{10.22} & \textbf{0.60} \\
\hline
\end{tabular}
\end{center}
\end{table}

Because we turn off the loop-closure detection function, the improvement of accuracy only comes from our 3D-SeqMOS algorithm. Compared with the original Lego, the absolute translation error is reduced by 5.061m, the start-stop position drift error is reduced by 4.83m, and the drift rate is reduced by 1.42\emph{\%}. Similarly, the experimental analysis based on Fast-Lio2 also reduces the errors by 4.83m, 14.07m and 0.82\emph{\%} respectively. Obviously, our method has great advantages in the positioning process of LiDAR-SLAM, and can effectively reduce the drift error in the real-time positioning process.

\subsection{Performance Improvement of LiDAR Mapping}
Our 3D-SeqMOS is main for robotics or autonomous driving tasks. In the previous sections, we analyzed the improvement effect of proposed 3D-SeqMOS on positioning and odometry. Next, we analyzed the improvement of LiDAR mapping. Mapping is an essential part of SLAM. High precision and pure LiDAR map is an important basis for path planning and navigation \cite{ravi2019lane}. When there are moving objects in the scene, the established global map will be mixed with many noise point clouds, which will affect the later navigation.
\begin{figure}[h]
\begin{center}
\subfloat[]{
\includegraphics[scale=0.36]{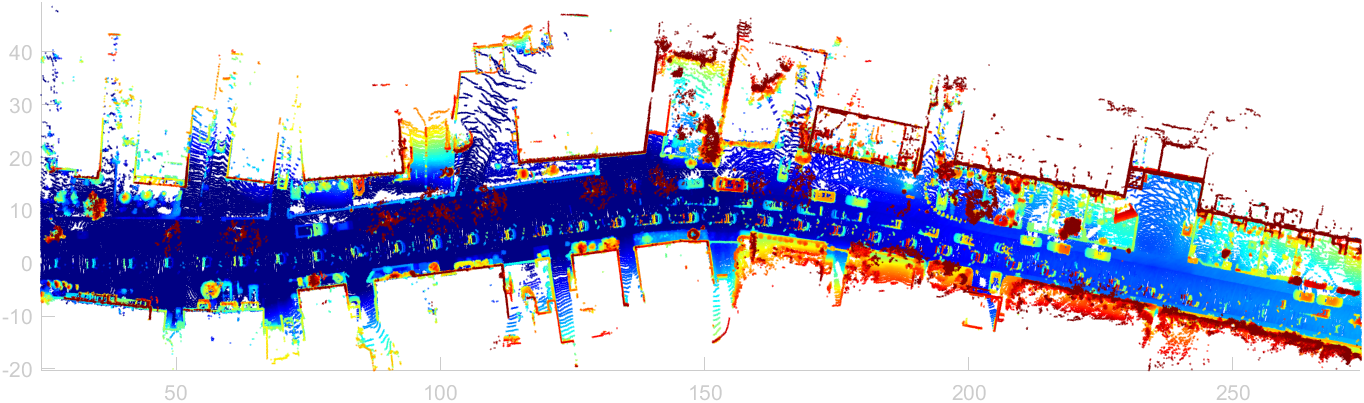}
\label{fig10.1}
}

\subfloat[]{
\includegraphics[scale=0.36]{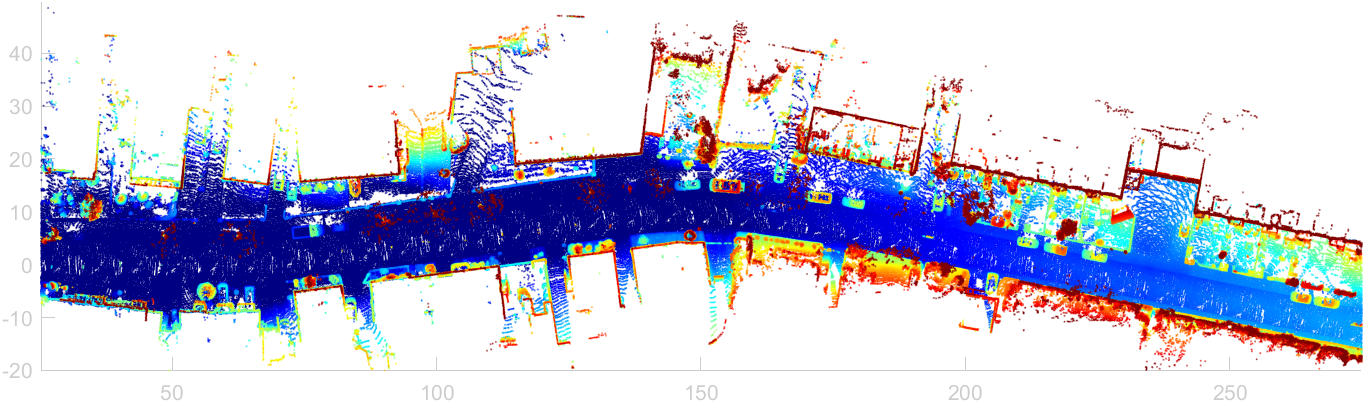}
\label{fig10.2}
}

\subfloat[]{
\includegraphics[scale=0.33]{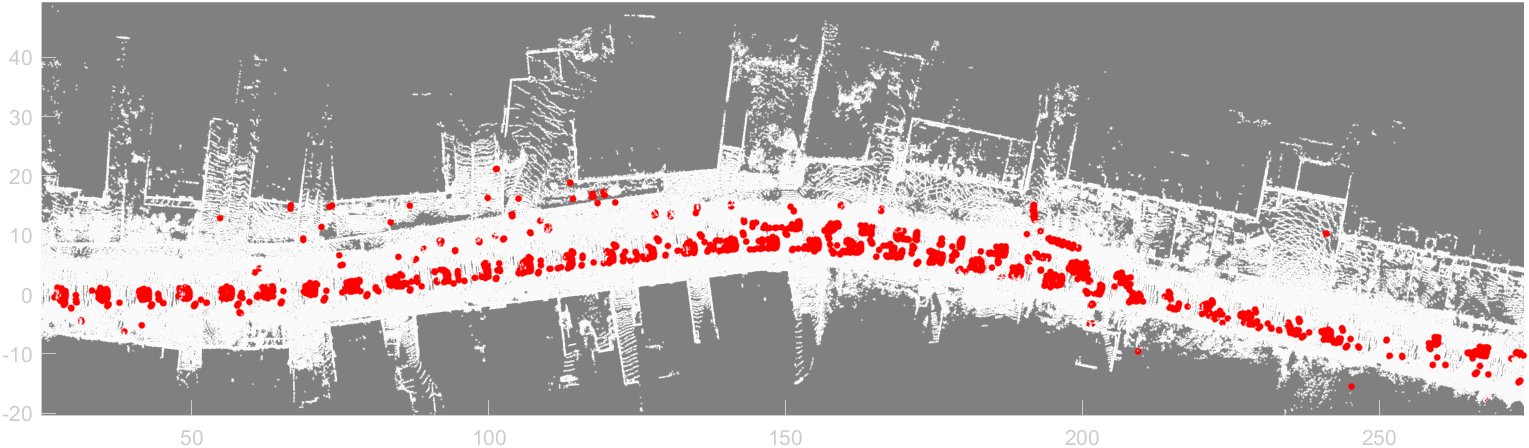}
\label{fig10.3}
}
\caption{Mapping performence results on Suquence 09, frame 1250-1450. (a) is the global map built with the raw point cloud, B is the result of filtering out moving objects based on our 3D-SeqMOS method, and (c) is the trajectory of moving objects in the global map (red mark).}
\label{fig10}
\end{center}
\end{figure}

\begin{figure}[h]
\begin{center}
\subfloat[]{
\includegraphics[scale=0.32]{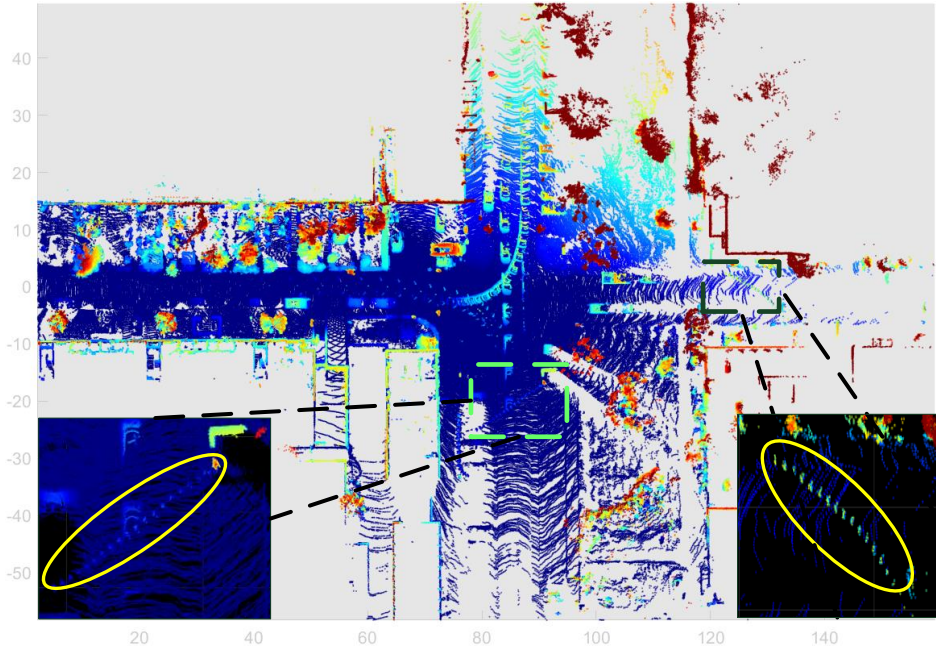}
\label{fig11.1}
}

\subfloat[]{
\includegraphics[scale=0.4]{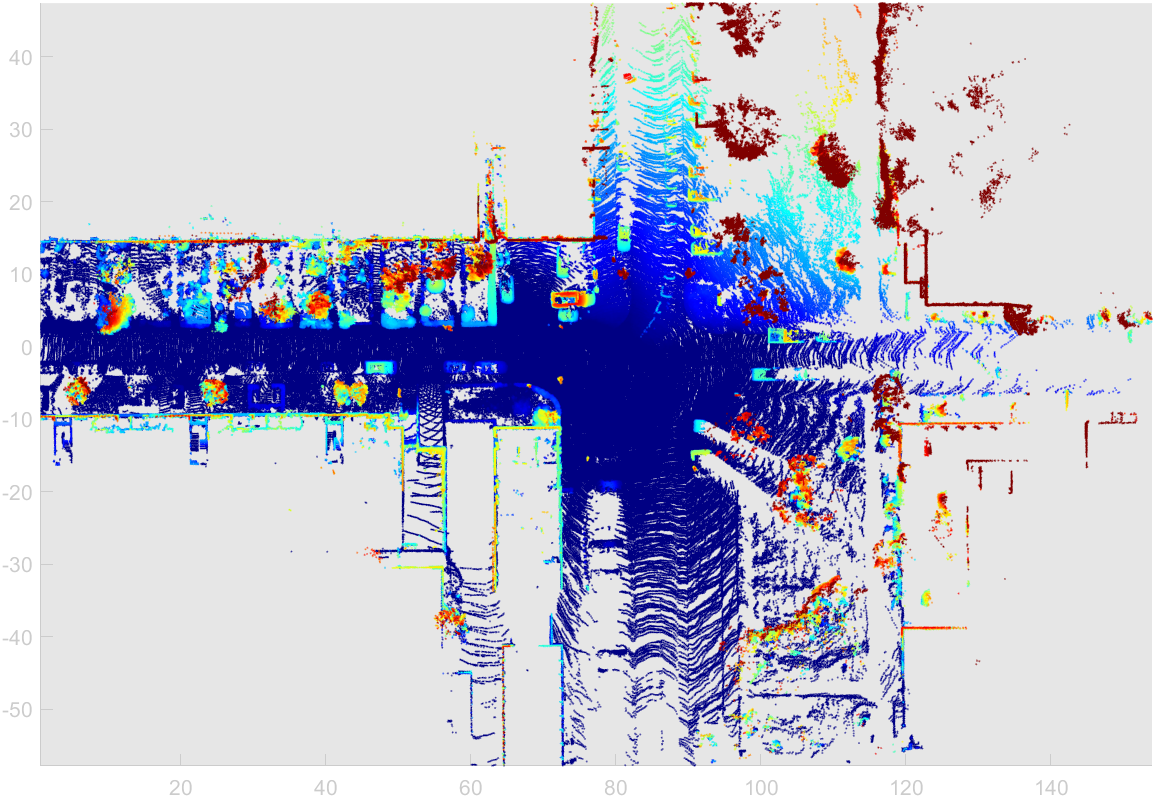}
\label{fig11.2}
}

\subfloat[]{
\includegraphics[scale=0.436]{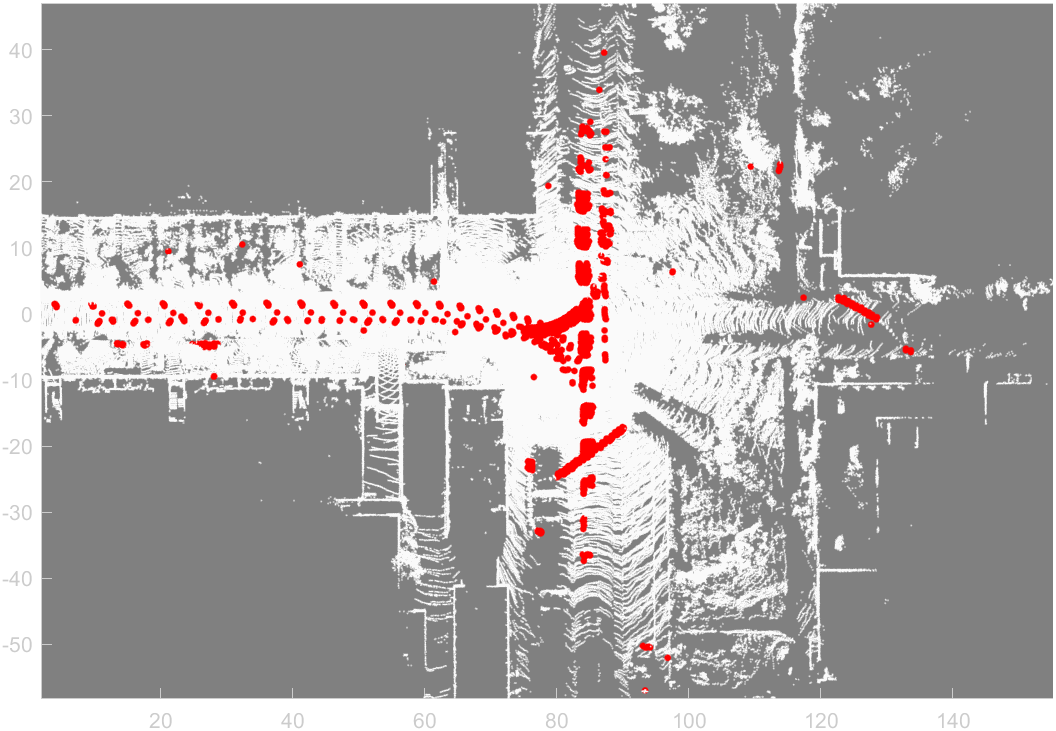}
\label{fig11.3}
}
\caption{Mapping performence results on Suquence 08, frame 3900-4050. (a) is the global map built with the raw point cloud, B is the result of filtering out moving objects based on our 3D-SeqMOS method, and (c) is the trajectory of moving objects in the global map (red mark).}
\label{fig11}
\end{center}
\end{figure}

To verify the effectiveness of our 3D-SeqMOS, we selected different scenes some continuous scans in sequence 08 (frames: 3900-4050) and sequence 09 (frames: 1250-1450) datasets respectively to establish a small-scale global map. It should be noted that these continuous scans contain more moving objects. As shown in Fig. \ref{fig10} and Fig. \ref{fig11}, the above part is a small-scale global map established based on the raw point cloud. We can see that in the road of Fig. \ref{fig10.1}, the motion trajectories of two moving objects run through the whole road, and in the yellow box of Fig. \ref{fig11.1}, in addition to moving cars, there are two pedestrians crossing the road. The motion trajectories of these moving objects are saved in the global map, which will cause great interference for path planning and navigation tasks. The both figures in the middle are based on the small-scale global map established with our 3D-SeqMOS. Through comparison, we can clearly see that the our established global map is very clean on the road, and there is almost no residual point cloud of moving objects. At the bottom of both figures are the moving object point cloud segmentation (red mark) in the global map. We can see that it occupies almost the whole road, and the road data can be considered as the most important constraint information in navigation. Therefore, we can intuitively find the advantages of our 3D-SeqMOS for mapping in autonomous driving.

\subsection{Kitti Odometry Benchmark}

From the previous experimental analysis, we can see that our 3D-SeqMOS method has a great improvement effect on SLAM, including positioning and mapping. Therefore, based on our 3D-SeqMOS, we also build a real time SLAM framework as described in III.G, and propose a odometry benchmark. In this section, our experiment aims to prove that our LiDAR odometry method has very competitive positioning accuracy. On several challenging sequences of KITTI dataset, we compare our positioning results with the more advanced mainstream SLAM framework, which can represent the research progress at current stage. In addition, we use absolute trajectory error (\emph{ATE}) as the evaluation criterion of our results with other exist methods.

\begin{table}[h]
\begin{center}
\caption{The positioning results based on absolute trajectory error on KITTI dataset. The black bold is the best.}
\label{table:3}
\begin{tabular}{llllllll}
\hline\noalign{\smallskip}
{\upshape Method} & 00 & 02 & 05 & 06 & 07 & 08 & 09\\
\noalign{\smallskip}
\hline
\noalign{\smallskip}
A-loam & 7.13 &	69.1 & 2.70 & 3.81 & 3.91 &	3.03 & 7.01\\
Lego-loam &	3.43 &	12.1 &	2.48 &	1.44 &	0.95 &	2.24 &	10.3 \\
Fast-Lio2 &	7.34 &	7.99 &	1.94 &	4.85 &	1.04 &	3.67 &	7.15 \\
ISC-loam &	1.60 &	4.77 &	2.49 &	1.03 &	0.56 &	4.88 &	2.31 \\
Suma &	1.14 &	44.1 &	0.86 &	0.68 &	0.4	 & 2.09 & 3.88 \\
Suma++ & 1.17 &	13.0 &	\textbf{0.64} &	0.56 &	0.37 &	2.44 &	1.19 \\
Sa-loam & \textbf{0.99} & 9.71 &	0.75 &	0.64 &	\textbf{0.36} &	3.24 &	1.2 \\
\textbf{Ours} &	1.20 &	\textbf{4.17} &	0.70 &	\textbf{0.53} &	0.43 &	\textbf{1.77} &	\textbf{1.12} \\
\hline
\end{tabular}
\end{center}
\end{table}

We compare our method with several advanced LiDAR-SLAM methods in TABLE \ref{table:3}. Since some codes are not available, we directly quote published paper results, including ISC-loam \cite{wang2020intensity}, Suma, Suma++, and Sa-loam \cite{li2021sa}, while the rest results are obtained by using its open source code and recommended parameters. Through the quantitative analysis of positioning accuracy, we can see that our proposed end-to-end LiDAR-SLAM framework based on our 3D-SeqMOS can achieve the state-of-the-art performance in most sequences.

\subsection{Performance Improvement of LiDAR Loop-Closure}
In SLAM system, back-end loop-closure detection is also an essential module. In this section, our experiment is to verify the improvement effect of 3D-SeqMOS for loop-closure detection. Since our method is based on semantic segmentation, we selected SG \cite{kong2020semantic}, an existing loop-closure detection method based on semantic segmentation, as the baseline. To quantify our performance, we use the precision-recall curve as the evaluation criterion. We only use our 3D-SeqMOS to optimize SG, and other parameters and model structures are all consistent with the original SG method, such as, we also take advantage of 1-fold train strategy, so the performance improvement can only come from our 3D-SeqMOS.

\begin{figure}[h]
\begin{center}
\subfloat[]{
\includegraphics[scale=0.45]{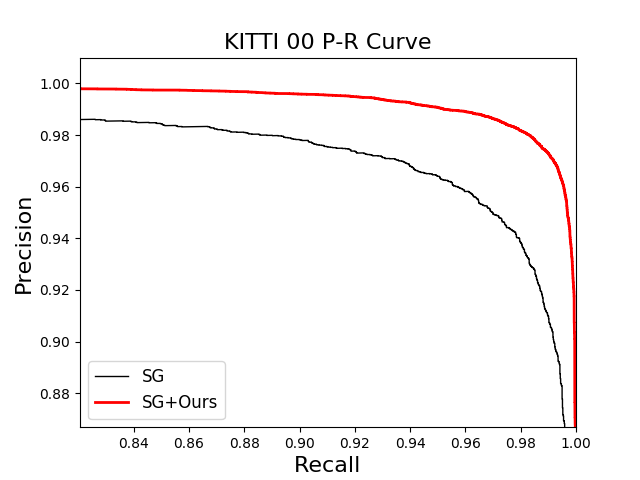}
\label{fig12.1}
}

\subfloat[]{
\includegraphics[scale=0.45]{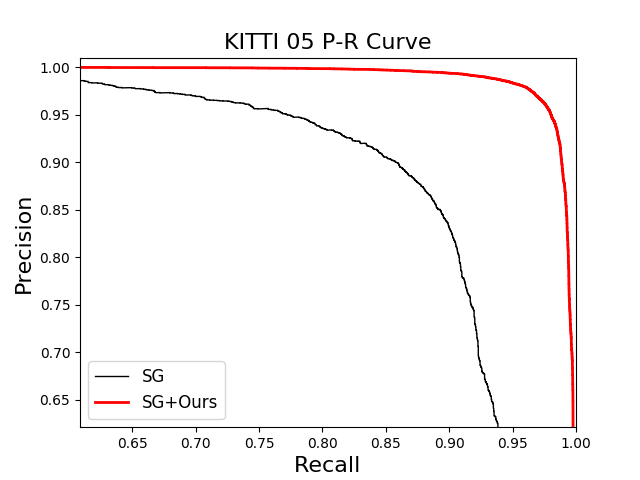}
\label{fig12.2}
}
\caption{The precision-recall curve results based on SG. (a) is the result on KITTI-00, (b) is the result on KITTI-05. Black is the orinal SG, and red is the SG with our 3D-SeqMOS.}
\label{fig12}
\end{center}
\end{figure}

We selected two KITTI sequences 00 and 05, which include loop-closure. As the precision-recall (P-R) curve shown in Fig. \ref{fig12}, we can see that black is the original SG-PR method and red is our optimized one. It is obvious from the P-R curve that our method has significantly improved the performance of the existing loop-closure detection method. Besides, to fully verify the effectiveness of our method, we selected all KITTI sequences with loop-closure, including 00, 02, 05, 06, 07 and 08. We calculate the maximum value of $F_1$ score to evaluate all P-R curves. The $F_1$ score is defined as:

\begin{equation}
    F_1 = 2 \times \frac{P\times R}{P+R}
\end{equation}
where $P$ denotes precision and $R$ denotes recall.

\setlength{\tabcolsep}{4pt}
\begin{table}[h]
\begin{center}
\caption{$F_1$ max scores on KITTI dataset.}
\label{table:4}
\begin{tabular}{llllllll}
\hline\noalign{\smallskip}
{\upshape Method} & 00 & 02 & 05 & 06 & 07 & 08 & {\upshape Mean}\\
\noalign{\smallskip}
\hline
\noalign{\smallskip}
SG &	0.969 &	0.891 &	0.905 &	0.971 &	0.967 &	0.900 &	0.934 \\
\textbf{{\upshape SG + Ours}} &	\textbf{0.982} & \textbf{0.971} &	\textbf{0.965} &	\textbf{0.987} &	\textbf{0.993}	 & \textbf{0.976} &	\textbf{0.979} \\
\hline
\end{tabular}
\end{center}
\end{table}
\setlength{\tabcolsep}{1.4pt}

It can be seen from $F_1$ values by TABLE \ref{table:4}  that our method improves the performance of SG on all loop-closure sequences. By filtering out the moving objects in the point cloud scene, not only in the point cloud registration, but also loop-closure detection by constructing semantic map, the SLAM system accuracy and stability can be improved. Thus, our 3D-SeqMOS has a great performance improvement for SLAM back-end loop-closure detection with existing mothods.

\section{Conclusion}
As for the moving object segmentation in autonomous driving and robotics, we designed a novel moving object segmentation network with the direct process of raw 3D point clouds instead of 2D images. To make better use of spatio-temporal information for moving segmentation, we designed a point cloud residual mechanism. The experimental results show that compared with the existing moving target segmentation methods, our method has obvious performance improvement and achieves the \emph{SOTA} performance. Based on our 3D-SeqMOS, we developed an end-to-end LiDAR-SLAM framework, including our moving object segmentation module and LiDAR odometry, and further verified the advantages of our method through KITTI dataset. Then, by integrating our 3D-SeqMOS with the existing SLAM methods, it showed the improvement effectiveness, including odometry, mapping, and loop-closure detection. To the best of our knowledge, 3D-SeqMOS may be the first moving object segmentation method using the sequential information of the raw 3D point clouds. Besides, in autonomous driving or intelligent robotics, our 3D-SeqMOS method can be well integrated with other existing methods and improve the performance, including odometry, mapping, loop-closure detection, and so on.

\section*{Acknowledgments}
This work was supported by Excellent Youth Foundation of Hubei Scientific Committee (2021CFA040), GuangDong Basic and Applied Basic Research Foundation (2021A1515110343), and Special Fund of Anhui University of Science and Technology (SKLMRDPC21KF20). The authors would also like to thank the reviewers for their insightful comments and suggestions.

\bibliographystyle{IEEEtran}
\bibliography{main}
\begin{IEEEbiography}[{\includegraphics[width=1in,height=1.25in,clip,keepaspectratio]{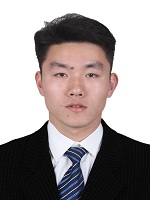}}]{Qipeng Li}
received the B.E. degree in automation from Huazhong Agricultural University,in 2016, and the M.S. degree in control science and engineering from Shandong University, in 2019, where he is currently pursuing the Ph.D. degree in the Sensing, Navigation \& Artificial Intelligence Lab \emph{(SNAIL)} at the State Key Laboratory of Information Engineering in Surveying, Mapping and Remote Sensing, Wuhan University. His research interests include intelligent robotics vision, point cloud processing, LiDAR-SLAM, positioning and navigation.
\end{IEEEbiography}

\begin{IEEEbiography}[{\includegraphics[width=1in,height=1.25in,clip,keepaspectratio]{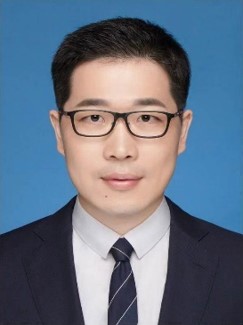}}]{Yuan Zhuang}
is a professor and the founder of the Sensing, Navigation \& Artificial Intelligence Lab \emph{(SNAIL)} at the State Key Laboratory of Information Engineering in Surveying, Mapping and Remote Sensing, Wuhan University, China. He received the Ph.D. degree in geomatics engineering from the University of Calgary, Canada in 2015. His current research interests include multi-sensors integration, realtime location system, wireless localization, Internet of Things (IoT), and machine learning for navigation applications. To date, he has co-authored over 100 academic papers and over 20 patents and has received over 10 academic awards. He is on the editorial board of Satellite Navigation and IEEE Access, the guest editor of the IEEE Internet of Things Journal, and a reviewer of over 20 IEEE journals.
\end{IEEEbiography}

\begin{IEEEbiography}[{\includegraphics[width=1in,height=1.25in,clip,keepaspectratio]{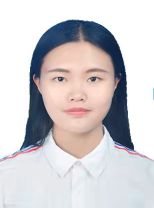}}]{Yiwen Chen}
received  the B.E. degree in automation and the M.S. degree in pattern recognition and intelligent system from Kunming University of Science and Technology in 2018 and 2021, respectively, where she is currently pursuing the Ph.D. degree in the Sensing, Navigation Artificial Intelligence Lab (SNAIL) at the State Key Laboratory of Information Engineering in Surveying, Mapping and Remote Sensing, Wuhan University. Her research focuses on intelligent robotics vision, point cloud processing, LiDAR-SLAM, positioning and navigation.
\end{IEEEbiography}

\begin{IEEEbiography}[{\includegraphics[width=1in,height=1.25in,clip,keepaspectratio]{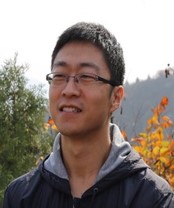}}]{Jianzhu Huai}
is a postdoctoral researcher at the State Key Laboratory of Surveying, Mapping and Remote Sensing, Wuhan University, China. Previously, he obtained a Ph.D. in Geodetic Engineering from The Ohio State University, USA, in 2017 for the work on collaborative mapping by using camera and/or IMU data collected from smartphones. The Android and iOS apps for data acquisition, MARS logger, has been open sourced. Then he worked 2.5 years at the Autonomous Personal Robot group in Segway Robotics where he developed localization, mapping, and calibration programs for sensor modalities including wheel encoder, camera, IMU, and lidar, prior to joining the state key lab. One of his completed open-source projects has extended the camera-IMU calibration package kalibr to deal with rolling shutter effect and noise identification. As a member of the Institute of Navigation (ION) since 2015, he has been investigating problems in state estimation, area mapping, and sensor calibration, aiming at cost-effective automation systems.
\end{IEEEbiography}

\begin{IEEEbiography}[{\includegraphics[width=1in,height=1.25in,clip,keepaspectratio]{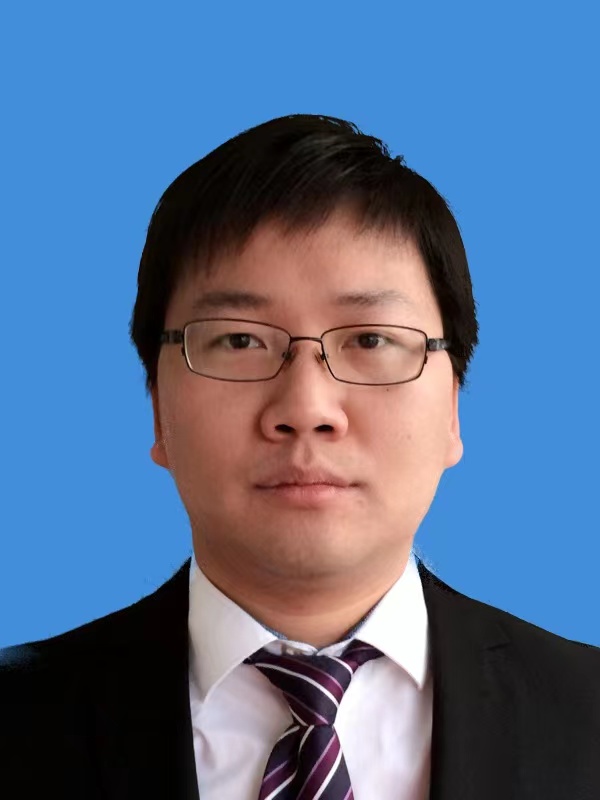}}]{Miao Li}
is an associate professor at the Institute of Technological Sciences, at Wuhan University, China. He received the Bachelor and Master’s degree in Mechanical Engineering from Huazhong University of Science and Technology (HUST), China in 2008 and 2011 respectively, and PhD in robotics from École Polytechnique Fédérale de Lausanne (EPFL) in Switzerland, 2016. He received the 2018 EPFL ABB Award for his PhD thesis titled Dynamic Grasp Adaptation – From Humans to Robots. His research focused on the design of intelligent system for advanced manufacturing and medical applications, especially on robot imitation learning, object grasping and manipulation, human-robot interaction.
\end{IEEEbiography}

\begin{IEEEbiography}[{\includegraphics[width=1in,height=1.25in,clip,keepaspectratio]{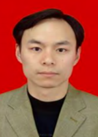}}]{Tianbing Ma}
(1981-), male, Ph.D., professor, received his bachelor's and master's degrees from Anhui University of Science and Technology in 2002 and 2005, and received his Ph.D. from Nanjing University of Aeronautics and Astronautics in 2014, and is now the deputy director of the State Key Laboratory of Deep Coal Mine Mining Response and Disaster Prevention and Control, mainly engaged in fault diagnosis, machine vision and vibration control research.
\end{IEEEbiography}


\begin{IEEEbiography}[{\includegraphics[width=1in,height=1.25in,clip,keepaspectratio]{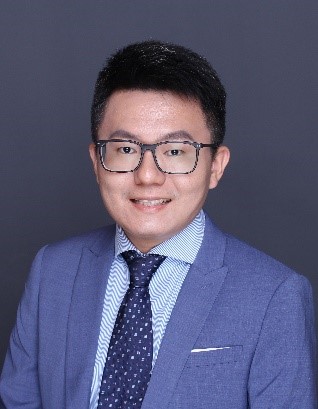}}]{Yufei Tang}
received the Ph.D. degree in Electrical Engineering from the University of Rhode Island, Kingston, RI, USA, in 2016. He is currently an Associate Professor with the Department of Electrical Engineering and Computer Science, and a Faculty Fellow with the Institute for Sensing and Embedded Network Systems Engineering, Florida Atlantic University, Boca Raton, FL, USA. His research interests include machine learning, data mining, dynamical systems, and renewable energy. Dr. Tang was a recipient of the IEEE International Conference on Communications Best Paper Award in 2014, the National Academies Gulf Research Program Early-Career Research Fellowship in 2019, and the National Science Foundation CAREER Award in 2022.
\end{IEEEbiography}

\begin{IEEEbiography}[{\includegraphics[width=1in,height=1.25in,clip,keepaspectratio]{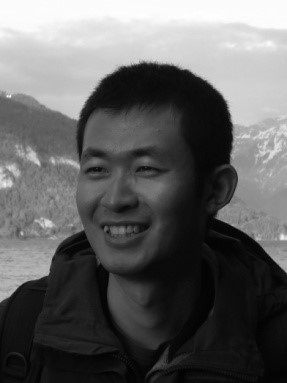}}]{Xinlian Liang}
 received his doctoral degree from Aalto University, Espoo, Finland, in 2013. He is a Prof. of Geoinformatics in Wuhan University (WH), China. Before joining WH, he was a Research Manager and the leader of the remote sensing of forest group with the Department of Remote Sensing and Photogrammetry, Finnish Geospatial Research Institute. His current research interests include the innovative geospatial techniques in modeling forest ecosystem, from all kinds of point clouds as well as imagery technologies.
\end{IEEEbiography}

\newpage




\end{document}